\documentclass[10pt,twocolumn,letterpaper]{article}

\usepackage{cvpr}              

\usepackage{graphicx} 
\usepackage{subcaption}
\usepackage{multirow}
\usepackage[accsupp]{axessibility} 
\usepackage{amsthm}
\usepackage{tabularx}
\usepackage{bbm}
\usepackage{amsmath}
\usepackage{algorithmic}
\usepackage{algorithm}
\usepackage{cuted}
\usepackage{amsmath,amssymb,amsthm}
\theoremstyle{definition}
\newtheorem{assumption}{Assumption}
\theoremstyle{plain}

\usepackage{capt-of}
\theoremstyle{plain}
\newtheorem{proposition}{Proposition}
\usepackage{booktabs}

\usepackage{makecell}
\usepackage{xcolor}
\definecolor{tocblue}{RGB}{25,80,165}
\usepackage[table]{xcolor}  
\definecolor{cvprblue}{rgb}{0.21,0.49,0.74}
\usepackage[pagebackref,breaklinks,colorlinks,allcolors=cvprblue]{hyperref}

\def\confName{CVPR}
\def\confYear{2027}

\title{From Concepts to Interactions: Dynamic Graph Semantic Recomposition for Safe Visual Generation}

\author{
Qinghui Gong$^{1}$ \quad
Jinshan Lai$^{2}$ \quad
Xunlei Chen$^{2}$ \\
Yu-Xuan Zhang$^{1}$ \quad
Hua Meng$^{1}$ \quad
Zhengchun Zhou$^{1,*}$ \\
\\
$^{1}$School of Information Science and Technology, Southwest Jiaotong University \\
$^{2}$School of Information and Software Engineering, \\
University of Electronic Science and Technology of China \\
Chengdu, China \\
\\
}

\begin{document}
\maketitle

\begin{strip}
    \centering
    \includegraphics[width=0.9\textwidth]{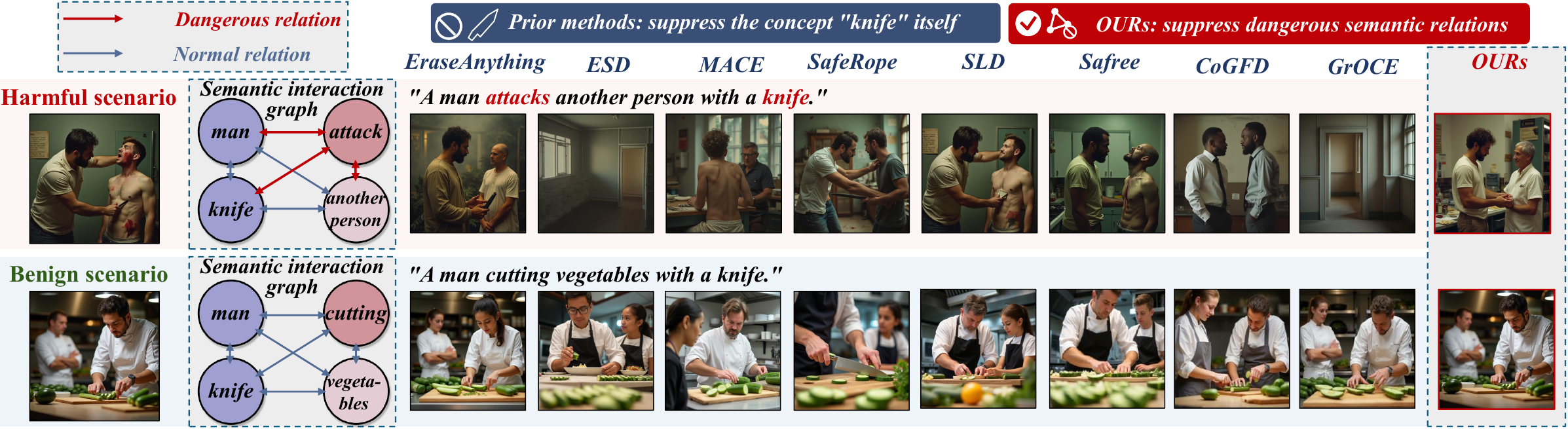}
    \captionof{figure}{
    \textbf{Motivation for relation-level safe generation.}
  Whether a concept is harmful depends on its contextual semantic relations. Prior erasure methods suppress “knife” unconditionally, leading to unrealistic outputs (e.g., cutting vegetables bare-handed in cooking scenes). DGSR selectively suppresses harmful semantic interactions and preserves benign, valid object usage.
    }
    \label{fig:motivation}
\end{strip}

\begin{abstract}
Recent advances in flow-matching Transformers have greatly improved Text-to-Image (T2I) and Text-to-Video (T2V) generation, but unconstrained generation risks producing illegal or unethical content. Safety alignment must therefore suppress unsafe content while preserving legitimate generation. A concept's safety depends on context: a knife used for cooking is safe, whereas its use in violence is harmful. Yet most concept erasure methods indiscriminately suppress static, isolated representations, overlooking evolving semantic interactions. Consequently, safe scenes containing potentially sensitive terms may be unintentionally suppressed, compromising generation quality. To address this, we propose Dynamic Graph Semantic Recomposition (DGSR), which dynamically constructs and updates a semantic interaction graph from evolving text visual states during sampling. Context aware graph reasoning captures semantic dependencies, while paired safe/unsafe supervision distinguishes harmful from legitimate interactions, enabling precise identification of and selective intervention on high risk semantic relationships while preserving benign semantics. DGSR freezes the pretrained backbone and supports both image and video generation. Experiments show that DGSR achieves the lowest Attack Success Rate (ASR) in 10 of 12 T2I evaluations and reduces the average T2V unsafe frame rate by 41.2\% relative to the strongest baseline, while maintaining high generation fidelity and temporal consistency. 
\end{abstract}

\section{Introduction}
\label{sec:introduction}

Rapid advances in diffusion models and flow-matching Transformer architectures~\cite{podell2023sdxl,flux2024} have  improved the visual quality and text alignment of  T2I and T2V generation~\cite{rombach2022high,esser2024scaling,wan2025wan}. However, these enhanced modeling capabilities also amplify the risk of generating unsafe content, such as nudity and violence~\cite{schramowski2023safe,yoon2025safree}. Effectively suppressing harmful generation while preserving benign semantics and general generative capabilities has therefore become a central challenge in controllable and safe visual generation.

Existing erasing methods fall into two categories: training-based and training-free approaches. Training-based methods erase target concepts through parameter fine-tuning or auxiliary modules, as exemplified by ESD and EraseAnything~\cite{gandikota2023erasing,gao2025eraseanything,ye2025t2vunlearning}. Training-free methods achieve safety control through weight editing or inference-time intervention, with representative approaches including UCE and SAFREE~\cite{gandikota2024unified,xu2025videoeraser,yoon2025safree}. Early methods uniformly treat individual concepts as the units of control, performing erasure or guidance based on static semantic representations and implicitly assuming that unsafety is an intrinsic, static property of a concept. Although recent methods such as CoGFD~\cite{nie2025cogfd}, GrOCE~\cite{han2026groce} and SafeRoPE~\cite{yang2026saferope} have advanced toward relation-level control through concept logic graphs and risk-related attention modulation, respectively, they still rely on textual priors to define semantic relations without coupling them to the evolving intermediate visual states during generation. This limits their ability to capture semantic interactions governed by latent dynamics.

From the perspective of the generation process, flow-matching models update latent states along the sampling trajectory~\cite{lipman2023flow,esser2024scaling}. Even when the text condition remains fixed, the visual responses of semantic units and their associations evolve as generation proceeds. For T2V, visual evidence additionally incorporates spatiotemporal variation across video frames~\cite{wan2025wan}. Safety control must therefore consider not only which semantics are present, but also how they participate in interactions in the current state. As illustrated in Fig.~\ref{fig:motivation}, ``a chef cutting vegetables'' and ``a person attacking someone with a knife'' both involve people and knives, yet different actions and interaction targets distinguish benign from harmful scenarios. Their difference in risk lies not in the presence of a knife, but in how the associated semantics combine and materialize as specific interactions. Globally suppressing shared concepts may consequently impair their legitimate expression.

These observations indicate that the core of safety control lies not in suppressing isolated concepts, but in dynamically identifying and selectively intervening in interactions among semantic units. This motivates our central question: \emph{Can we dynamically identify emerging high-risk semantic relations from the current generation state and iteratively update intervention strategies as generation evolves?}

Addressing this question requires explicitly modeling the compositional dependencies and dynamic interactions among multiple semantic units in a given context, which are difficult to capture adequately through isolated concept representations. We therefore introduce a graph representation in which nodes represent semantic units and edges characterize their associations. Relation-conditioned message passing integrates local interaction evidence with contextual semantics, providing a structured basis for identifying high-risk relations and selectively intervening in them.

Building on these insights, we propose DGSR, which shifts safety control from static concept suppression to graph-based intervention in dynamic semantic relations. DGSR combines text conditions with current intermediate visual representations to construct a Dynamic Semantic Interaction Graph (DSIG): semantic units, such as entities and actions, serve as nodes, while candidate relation edges are constructed from their joint visual responses. Contextual reasoning aggregates neighborhood information to predict suppression strengths for individual semantic relations and maps these predictions to Transformer attention layers, enabling dynamic intervention as latent states are iteratively updated. The framework models semantic interactions using spatial image tokens and spatiotemporal video tokens, providing a unified approach to T2I and T2V generation. Experiments show that DGSR achieves the best performance on 10 of the 12 T2I ASR evaluations, with a 11.4\%  relative reduction in ASR for Violence compared with the strongest baseline. For T2V, DGSR reduces the average unsafe generation rate by
41.2\% relative to the per-setting strongest baselines, while
preserving generation fidelity and temporal consistency.

Our main contributions are summarized as follows:

\begin{itemize}
\item 
We investigate generation-state-dependent semantic interactions as the basis for fine-grained safety control, moving beyond isolated concept suppression.

\item 
We introduce DSIG, which dynamically constructs semantic nodes and relation edges from semantic spans and current text--visual grounding. Graph-based relation reasoning produces suppression signals for high-risk interactions, enabling generation state aware safety intervention.

\item
Our framework unifies graph-based semantic modeling and feedback intervention over spatial and spatiotemporal features, supporting safe T2I and T2V generation with flow-matching and diffusion models.

\end{itemize}

\section{Related Work}
\label{sec:related_work}


 \paragraph{Safe T2I/T2V Generation.}
Safe T2I and T2V generation methods mainly follow three strategies:
inference-time intervention, training-based erasure, and closed-form
weight editing.
Inference-time approaches such as SLD, SAFREE, and
VideoEraser~\cite{schramowski2023safe,yoon2025safree,xu2025videoeraser}
suppress unsafe content through safety guidance or representation
adjustment, while SafeRoPE~\cite{yang2026saferope} performs selective
attention modulation through risk-aware rotation.
Training-based methods optimize model parameters for concept erasure
in images~\cite{gandikota2023erasing,bui2024erasing,gao2025eraseanything}
and videos~\cite{liu2024unlearning,ye2025t2vunlearning,fan2026eraseanything},
with video-oriented designs considering non-target
semantic preservation and temporal consistency.
UCE, RECE, and ICE~\cite{gandikota2024unified,gong2025reliable,dasbiswas2026ice}
perform concept erasure through analytical weight updates,
with ICE supporting both image and video generation.
CoGFD~\cite{nie2025cogfd} introduces a concept logic graph
for compositional concept erasure.
DGSR constructs semantic graphs from evolving text--visual
representations and uses contextual relation reasoning to guide
generation-state-aware intervention in both image and video generation.


\paragraph{Unsafe Generation Formulation.}
We consider flow-matching generative models, such as FLUX,
which employ multimodal Transformers for visual generation.
Given a text embedding $e_p=\mathcal{T}(p)$, a timestep $t$,
and a latent state $x_t$, the pretrained network
$F_\theta(x_t,t,e_p)$ predicts the velocity field and
generates a visual sample $x=G_\theta(z,p)$ through iterative
sampling, where $\theta$ denotes the parameters
and $z\sim\mathcal{N}(0,I)$.
Unsafe generation may arise from harmful semantic interactions
induced by a prompt $p$. Let $r_{\mathrm{unsafe}}$ denote
such an interaction. Its occurrence probability is defined as
\begin{equation}
P_\theta(r_{\mathrm{unsafe}}\mid p)
=
\mathbb{E}_{x\sim P_\theta(x\mid p)}
\left[
\mathbbm{1}\{x\models r_{\mathrm{unsafe}}\}
\right],
\label{eq:unsafe_relation_prob}
\end{equation}
where $P_\theta(x\mid p)$ denotes the conditional generation
distribution, and $\mathbbm{1}\{x\models r_{\mathrm{unsafe}}\}$
indicates whether the generated sample contains the harmful
interaction.


\begin{figure*}
    \centering
    \includegraphics[width=0.95\linewidth]{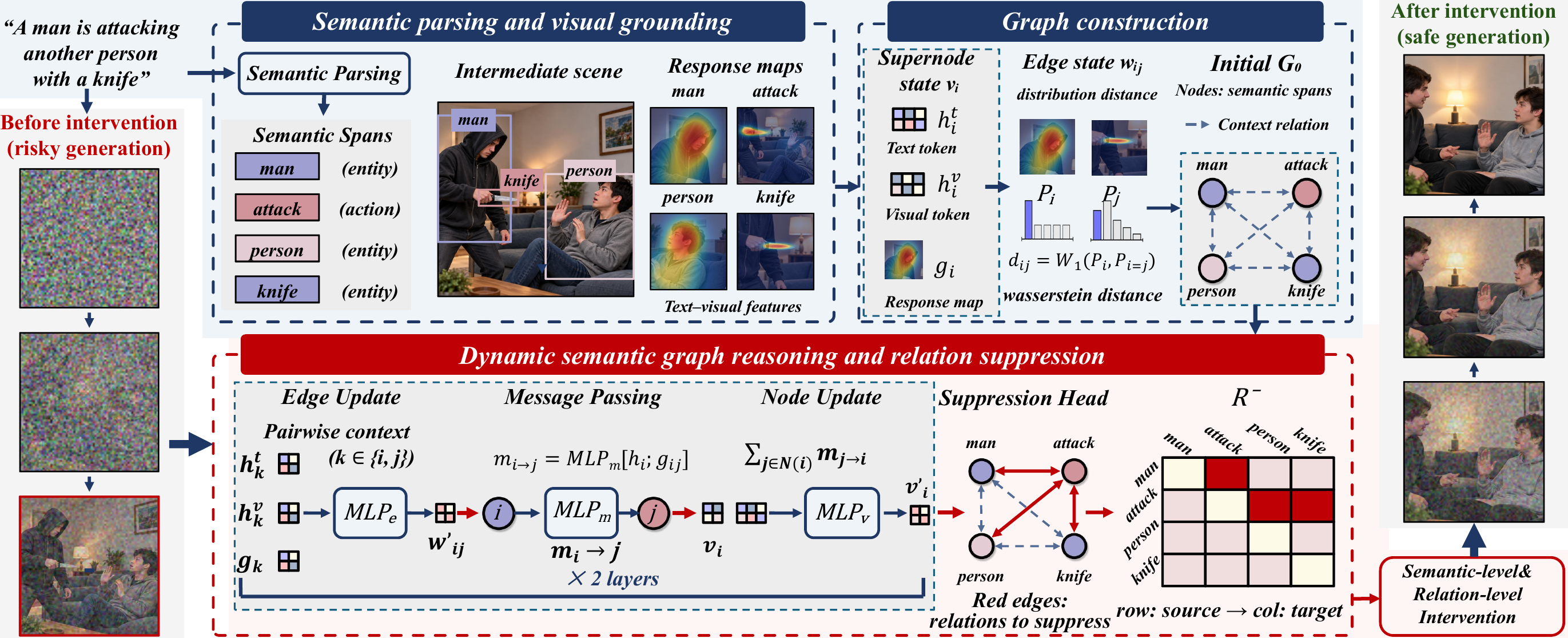} 
\caption{
\textbf{Overview of DGSR.}
DGSR constructs semantic supernodes from text--visual features
and establishes a dynamic semantic interaction graph (DSIG)
using Wasserstein-based grounding similarities.
A two-layer Edge-GNN captures contextual dependencies
and predicts harmful relation suppression strengths.
These signals are mapped back to Transformer attention
through semantic-level and relation-level recomposition,
enabling selective suppression of harmful interactions
while preserving benign semantics.
}
    \label{fig:framework}
\end{figure*}

\section{Method}
\label{sec:method}

\subsection{Framework Overview}
\label{sec:overview}

Risks in visual generation depend not only on the semantic units in the
text but also on their interactions in the current generation state.
DGSR explicitly models semantic relations using dynamic graphs
and learns risk patterns through supervision from paired safe/unsafe
samples, shifting safety control from isolated concept suppression to
interaction-level intervention. The framework
(Fig.~\ref{fig:framework}) consists of three steps:
\textbf{(1) Dynamic Semantic Interaction Graph Construction.}
We construct DSIG using semantic spans as nodes and joint visual responses
as evidence of association, converting current cross-modal interactions
into a structured graph representation.
\textbf{(2) Context-aware Relation Prediction.}
Relation-conditioned message passing integrates node, edge, and
neighborhood information to predict the suppression strength of each
directed relation, yielding the relation suppression matrix $\mathbf R^-$.
\textbf{(3) Inference-time Semantic Recomposition.}
We perform risk gating based on $\mathbf R^-$ and modulate Transformer
attention through semantic-level suppression and relation-level
recomposition.

\subsection{Dynamic Semantic Interaction Graph}
\label{sec:dsig}

To jointly represent semantic identity and the current visual state,
we construct a dynamic semantic interaction graph at sampling step $t$
and Transformer block $l$:
\begin{equation}
\mathcal G_t^l=(\mathcal V,\mathcal E,\mathbf X,\mathbf E_a),
\label{eq:graph_def}
\end{equation}
where $\mathcal V$ and $\mathcal E$ denote the sets of semantic nodes
and candidate directed relations, while $\mathbf X$ and
$\mathbf E_a$ are the node features and edge attributes in the current
generation state. 

\noindent\textbf{Semantic Grounding and Node States.}
Individual sub-tokens may not fully capture the semantics of an entity
or action, while fixed textual representations alone cannot reflect
their current visual manifestation. We therefore group the text
sub-tokens into $N$ semantic spans, with $v_i$ denoting the $i$-th
semantic node and $\mathcal S_i$ its set of text token indices.
To associate each node with the current visual state, we derive dynamic
grounding from text--visual attention within the Transformer.
Suppose the current layer contains $M$ visual tokens and $H$ attention
heads, each with dimension $D_h$. We average the text keys of span $i$
within head $h$ to obtain $\bar{\mathbf K}_{i,h}$ and compute its
affinity with the visual query $\mathbf Q_{p,h}$:
\begin{equation}
\begin{aligned}
Z_{p,i}&=\frac{1}{H\sqrt{D_h}}\sum_{h=1}^{H}
\mathbf Q_{p,h}^{\top}\bar{\mathbf K}_{i,h},&
A_{p,i}=\operatorname{softmax}_{i}(Z_{p,i}).
\end{aligned}
\label{eq:semantic_affinity}
\end{equation}
where $p\in\{1,\ldots,M\}$ indexes visual tokens, and
$\mathbf Z,\mathbf A\in\mathbb R^{M\times N}$ denote the cross-modal
matching scores and the relative responses of visual tokens to semantic
spans. The span-level key aggregation is detailed in
Appendix~\ref{app:grounding}.

While $A_{p,i}$ describes the relative responses to different semantic
spans at a visual position, constructing nodes and semantic relations
also requires the response distribution of each span over the entire
visual domain. We therefore normalize over visual positions:
\begin{equation}
P_{p,i}=\frac{A_{p,i}}{\sum_{q=1}^{M}A_{q,i}},\qquad
\mathbf P_i=(P_{1,i},\ldots,P_{M,i})^{\top}.
\label{eq:visual_distribution}
\end{equation}
The resulting $\mathbf P_i$ represents the current visual grounding
distribution of semantic span $i$. We further aggregate its textual
identity, corresponding visual state, and response statistics into
a semantic supernode:
\begin{equation}
\begin{aligned}
\mathbf x_i^{\mathrm{text}}
&=\frac{1}{|\mathcal S_i|}
\sum_{k\in\mathcal S_i}\mathbf h_k^{\mathrm{text}},
\mathbf x_i^{\mathrm{visual}}
=\sum_{p=1}^{M}P_{p,i}\mathbf h_p^{\mathrm{visual}},\\
\mathbf x_i&=[\mathbf x_i^{\mathrm{text}};
\mathbf x_i^{\mathrm{visual}};\mathbf x_i^{\mathrm{stat}}].
\end{aligned}
\label{eq:node_feature}
\end{equation}
where $\mathbf h_k^{\mathrm{text}}$ and
$\mathbf h_p^{\mathrm{visual}}$ are the hidden states of text and
visual tokens, respectively, and $\mathbf x_i^{\mathrm{stat}}$
summarizes the maximum and mean responses and the grounding entropy
(Appendix~\ref{app:grounding}). Together, the nodes form
$\mathbf X=[\mathbf x_1;\ldots;\mathbf x_N]$, jointly representing
the semantic content and current visual state of each span.

\noindent\textbf{Semantic Relation Construction.}
Nodes describe the current states of individual semantic units, while
relation edges further characterize their associations in the visual
domain. We define the candidate directed relation set as
$\mathcal E=\{(i,j)\mid i\ne j,\,
i,j\in\{1,\ldots,N\}\}$.
At the same sampling step $t$ and Transformer layer $l$, we use the
first-order Wasserstein distance to measure the difference between
the visual grounding distributions $\mathbf P_i,\mathbf P_j$ of
semantic nodes $i,j$ and convert it into an initial edge weight:
\begin{equation}
\begin{aligned}
d_{ij}&=\mathcal W_1(\mathbf P_i,\mathbf P_j;\mathbf C),
&w_{ij}=\exp(-d_{ij}/\tau_{\mathrm{ot}}).
\end{aligned}
\label{eq:edge_weight}
\end{equation}
where $\mathbf C$ is the transport cost matrix between visual token
positions, and $\tau_{\mathrm{ot}}>0$ is a distance scaling parameter.
The weight $w_{ij}\in(0,1]$ represents the visual association strength
in the current state: closer grounding distributions yield larger
edge weights. All edge attributes form
$\mathbf E_a=\{w_{ij}\mid(i,j)\in\mathcal E\}$.
The optimal transport formulation is provided in
Appendix~\ref{app:ot}.

\subsection{Context-aware Relation Prediction}
\label{sec:reasoning}

The initial edge weights provide evidence of visual associations.
We use a two-layer Edge-GNN to predict relation suppression strengths
by combining node semantics with graph context. We first project the
node and edge attributes into a shared hidden space:
\begin{equation}
\mathbf h_i^{(0)}=f_v(\mathbf x_i),\qquad
\mathbf g_{ij}^{(0)}=f_e(w_{ij}),
\label{eq:projection}
\end{equation}
where $f_v,f_e$ are the node and edge projection networks,
respectively, and $\mathbf h_i^{(0)},\mathbf g_{ij}^{(0)}$
are the initial hidden states.

To incorporate semantic evidence from both the endpoints and the
neighborhood of each relation, we alternate directed edge updates
with node updates through relation-conditioned message passing:
\begin{equation}
\begin{aligned}
\mathbf g_{ij}^{(q+1)}
&=\mathcal U_e^{(q)}(
\mathbf h_i^{(q)},\mathbf h_j^{(q)},\mathbf g_{ij}^{(q)}),\\
\mathbf h_j^{(q+1)}
&=\mathcal U_v^{(q)}\!\left(\mathbf h_j^{(q)},
\sum_{i\ne j}\mathcal M^{(q)}(
\mathbf h_i^{(q)},\mathbf g_{ij}^{(q+1)})\right).
\end{aligned}
\label{eq:context_reasoning}
\end{equation}
where $q\in\{0,1\}$ indexes reasoning rounds, and
$\mathbf h_i^{(q)},\mathbf g_{ij}^{(q)}$ denote the node and directed
edge states at round $q$. The functions
$\mathcal U_e^{(q)},\mathcal U_v^{(q)}$ update edge and node states,
respectively, while $\mathcal M^{(q)}$ is the relation-conditioned
message function. The update functions use residual MLPs and
LayerNorm, as detailed in Appendix~\ref{app:gnn}.

After two rounds of propagation, each directed edge obtains a state
$\mathbf g_{ij}^{(2)}$ that incorporates endpoint semantics and
neighborhood information. The relation suppression head further
combines this state with the updated node states to predict its
suppression strength:
\begin{equation}
R_{ij}^{-}=f_{\mathrm{sup}}\!\left(
[\mathbf h_i^{(2)};\mathbf h_j^{(2)};\mathbf g_{ij}^{(2)}]\right),
\quad R_{ij}^{-}\in[0,1].
\label{eq:relation_score}
\end{equation}
where $f_{\mathrm{sup}}$ is a relation suppression head with bounded
outputs, and $R_{ij}^-$ denotes the predicted suppression strength
for the directed relation $i\!\rightarrow\!j$ in the current state.
All non-self-loop scores form the relation suppression matrix
$\mathbf R^-\in[0,1]^{N\times N}$, with its diagonal entries set to zero.

\subsection{Inference-time Semantic Recomposition}
\label{sec:recomposition}

At each selected sampling step and Transformer block, DGSR reconstructs
the graph and predicts $\mathbf R^-$ to determine whether to intervene.
Once activated, it sequentially modulates the participation strengths
of semantic units and the attention contributions of semantic relations.

\noindent\textbf{Relation-based Risk Gating.}
To prevent a small number of high-risk interactions from being diluted
by many low-scoring relations, we use the mean of the top $10\%$ of
relation suppression scores as the gating score for the current graph.
Sorting the non-self-loop relation scores
$\{R_{ij}^-\}_{(i,j)\in\mathcal E}$ in descending order as
$r_{(1)}\ge\cdots\ge r_{(|\mathcal E|)}$, we compute
\begin{equation}
\begin{aligned}
k_g&=\max\!\left(1,\left\lceil0.1|\mathcal E|\right\rceil\right),
s_G=\frac{1}{k_g}\sum_{m=1}^{k_g}r_{(m)},\\
a_t^l&=\mathbbm{1}\{s_G\ge\tau_{\mathrm{gate}}\}.
\end{aligned}
\label{eq:risk_gate}
\end{equation}
where $|\mathcal E|=N(N-1)$, and $\tau_{\mathrm{gate}}$ is the activation threshold, with backbone-specific values provided in Appendix \ref{infer_impl}. When $a_t^l=0$, the original attention is preserved; otherwise, the full $\mathbf R^-$ is used for the two-level recomposition.

\noindent\textbf{Semantic-level Suppression.}
High-risk semantic interactions may involve multiple semantic
spans. To regulate their participation in attention while
avoiding excessive suppression caused by accumulating
low-risk relations, we define the suppression strength
of each semantic node based on its highest-risk
outgoing relation:
\begin{equation}
\begin{aligned}
c_i &= \max_{j\ne i} R_{ij}^-,&
\widetilde{\mathbf s}_k
= (1-c_i)\mathbf s_k,
\qquad k\in\mathcal S_i.
\end{aligned}
\label{eq:token_recomposition}
\end{equation}
where $c_i\in[0,1]$ denotes the semantic-level
suppression strength of span $i$, determined by its
strongest outgoing risk relation.
$\mathbf s_k$ and $\widetilde{\mathbf s}_k$ represent
the original and adjusted states of text token $k$.

\noindent\textbf{Relation-level Recomposition.}
Building on semantic-level modulation, we further preserve the directed
relation structure in the full suppression matrix $\mathbf R^-$ and
combine it with the current visual grounding to map relation-level
decisions to position-dependent visual--semantic suppression:
\begin{equation}
\mathbf S=\mathbf A(\mathbf R^-)^{\top},\qquad
\widetilde{\mathbf A}=\operatorname{softmax}_{i}(\mathbf Z-\mathbf S),
\label{eq:relation_recomposition}
\end{equation}
where $\mathbf A,\mathbf Z$ are the semantic responses and matching
scores from graph construction, and $\mathbf S$ represents
position-dependent relation suppression.
Elementwise, $S_{p,i}=\sum_j A_{p,j}R^-_{ij}$:
the suppression score of relation $i\!\rightarrow\!j$ modulates
the contribution of span $i$ at position $p$ through the visual response
of context span $j$. We then use span-level values to convert the
semantic response change $\widetilde{\mathbf A}-\mathbf A$ into
an increment to the visual attention output:
\begin{equation}
\begin{aligned}
\Delta_{\mathrm{sem}}&=
(\widetilde{\mathbf A}-\mathbf A)\mathbf V_{\mathrm{span}},&
\widetilde{\mathbf y}_I=\mathbf y_I+\Delta_{\mathrm{sem}}.
\end{aligned}
\label{eq:semantic_message}
\end{equation}
where $\mathbf V_{\mathrm{span}}\in\mathbb R^{N\times D_v}$ is the
span-level value matrix, $\mathbf y_I$ is the visual attention output
after semantic-level suppression, $\widetilde{\mathbf y}_I$
is the output after adding the relation-level increment. The construction of $\mathbf V_{\mathrm{span}}$
is detailed in Appendix~\ref{app:grounding}.

\subsection{Progressive Risk-aware Training}
\label{sec:training}

Having described graph construction and intervention in DGSR,
we now present its training strategy. We freeze the backbone $F_\theta$
and train the auxiliary module in three stages: first learning risk
representations from graph-level labels, then supervising relation
suppression scores with paired safe/unsafe conditions, and finally
jointly optimizing generation alignment and benign preservation losses.

\noindent\textbf{Graph-level Risk Learning.}
We first learn risk-related graph representations from unsafe/benign
graph-level labels by introducing an auxiliary classification head
on the Edge-GNN and minimizing
\begin{equation}
\hat b=f_\psi(\mathcal G_t^l),\qquad
\mathcal L_{\mathrm{graph}}=\operatorname{BCE}(\hat b,b).
\label{eq:graph_loss}
\end{equation}
where $b\in\{0,1\}$ is the graph-level label ($1$ indicates unsafe),
$\hat b\in[0,1]$ is the predicted risk probability, and
$\operatorname{BCE}$ denotes binary cross-entropy.
The network $f_\psi$ comprises an Edge-GNN encoder with parameters
$\psi_g$ and an auxiliary classification head with parameters $\psi_c$,
where $\psi=\{\psi_g,\psi_c\}$. The encoder learned at this stage is
used for subsequent relation prediction, while the auxiliary
classification head is used only for training supervision.

\noindent\textbf{Counterfactual Relation Supervision.}
Given paired unsafe/safe conditions $(c_u,c_s)$, we construct two graphs
at the same latent state $\mathbf x_t$, timestep $t$, and layer $l$,
and obtain $\mathbf R_u^-$ and $\mathbf R_s^-$ using a shared relation
prediction network. Based on the target actions modified in the paired
texts and their participating entities, we identify the respective
candidate relation sets $\mathcal E_u^f$ and $\mathcal E_s^f$
and aggregate the corresponding suppression scores:
\begin{equation}
s_q =
\frac{1}{|\mathcal{E}_q^f|}
\sum_{(i,j)\in\mathcal{E}_q^f}
R_{q,ij}^-,
\quad q\in\{u,s\}.
\label{eq:relation_target}
\end{equation}
where $s_u$ and $s_s$ denote the aggregated risk scores for unsafe
and benign interactions, respectively. To improve risk discrimination
and avoid erroneous suppression, we use a relation contrastive loss:
\begin{equation}
\begin{aligned}
\mathcal{L}_{\mathrm{relation}}
={}[m-s_u+s_s]_+
+\operatorname{BCE}(s_u,1)
+\operatorname{BCE}(s_s,0).
\end{aligned}
\label{eq:relation_loss}
\end{equation}
where $m>0$ is the contrastive margin. The construction of candidate
relations and the training configuration are detailed in
Appendix~\ref{app:counterfactual_supervision}.

\noindent\textbf{Generation Alignment and Preservation.}
Finally, we use the velocity prediction under the safe condition
at the same latent state as a reference to optimize the effect of
relation intervention on generation predictions. We also preserve
the original generation behavior by constraining prediction consistency
under benign conditions.
Let $\mathbf y_u=F_\theta(\mathbf x_t,t,c_u)$ and
$\mathbf y_s=F_\theta(\mathbf x_t,t,c_s)$ be the velocity predictions
of the frozen backbone under the paired conditions.
We define the target change and the actual change induced by
intervention as
\begin{equation}
\begin{aligned}
\Delta^*&=\mathbf y_s-\mathbf y_u,
\Delta=F_{\theta,\phi}^{\mathrm{DGSR}}(\mathbf x_t,t,c_u)-\mathbf y_u.
\end{aligned}
\label{eq:generation_delta}
\end{equation}
where $F_{\theta,\phi}^{\mathrm{DGSR}}$ denotes the velocity prediction
with DGSR attention intervention, and $\phi$ contains the learnable
parameters of the auxiliary module. To match both the magnitude
and direction of the changes, we use
\begin{equation}
\begin{aligned}
\mathcal L_{\mathrm{align}}
={}&\frac{\operatorname{MSE}(\Delta,\Delta^*)}
{\operatorname{Mean}((\Delta^*)^2)+\epsilon}
+1-\cos\!\left(\operatorname{vec}(\Delta),
\operatorname{vec}(\Delta^*)\right),
\end{aligned}
\label{eq:alignment_loss}
\end{equation}
where $\operatorname{MSE}$ denotes mean squared error,
$\operatorname{Mean}$ the mean over elements,
$\operatorname{vec}$ vectorization, and $\cos$ cosine similarity;
$\epsilon>0$ is a numerical stability constant.
For a benign text condition $c_b$, we further constrain the velocity
predictions before and after intervention to remain close:
\begin{equation}
\mathcal L_{\mathrm{retain}}=
\operatorname{MSE}\!\left(
F_{\theta,\phi}^{\mathrm{DGSR}}(\mathbf x_t,t,c_b),
F_\theta(\mathbf x_t,t,c_b)\right).
\label{eq:retain_loss}
\end{equation}
The third stage jointly minimizes
$\mathcal L_{\mathrm{III}}=
\lambda_{\mathrm{align}}\mathcal L_{\mathrm{align}}
+\lambda_{\mathrm{retain}}\mathcal L_{\mathrm{retain}}$,
where $\lambda_{\mathrm{align}}$ and $\lambda_{\mathrm{retain}}$
are loss weights.

\subsection{Understanding Relation-guided Intervention}
\label{sec:theoretical_analysis}

\paragraph{Relation-guided semantic reweighting.}
For the semantic recomposition in Eq.~\eqref{eq:relation_recomposition},
we analyze how relation suppression affects the ratios of semantic
responses.

\begin{assumption}[Finite semantic logits]
\label{ass:finite_logits}
For any visual position $p$ and semantic span $i$,
$Z_{p,i}$ is a finite real number.
\end{assumption}

\begin{proposition}[Relative semantic reweighting]
\label{thm:semantic_reweighting}
Under Assumption~\ref{ass:finite_logits}, any two semantic spans $i,j$
satisfy
\begin{equation}
\frac{\widetilde A_{p,i}}{\widetilde A_{p,j}}
=
\frac{A_{p,i}}{A_{p,j}}
\exp\!\left[-(S_{p,i}-S_{p,j})\right].
\label{eq:relative_reweighting}
\end{equation}
A larger relative penalty reduces the response ratio of the corresponding
span, whereas equal penalties leave the ratio unchanged.
The full proof is provided in Appendix~\ref{app:semantic_reweighting}.
\end{proposition}

\paragraph{Local alignment and sampling behavior.}
By the definitions of the increments in generation alignment,
\begin{equation}
\Delta-\Delta^*
=
F_{\theta,\phi}^{\mathrm{DGSR}}(x_t,t,c_u)
-
F_\theta(x_t,t,c_s).
\label{eq:alignment_equivalence}
\end{equation}
The normalized squared-error term in the alignment loss constrains
the difference between the intervened prediction and the safe-condition
reference prediction at the state.

Let $v_\phi,v_\star$ denote the intervention and reference fields above. On a shared time grid $t_k=t_0+kh$ that increases along
the sampling direction, consider the Euler updates:
\begin{equation}
\begin{aligned}
z_{k+1}&=z_k+h\,v_\phi(z_k,t_k),\\
\bar z_{k+1}&=\bar z_k+h\,v_\star(\bar z_k,t_k),
\end{aligned}
\label{eq:theory_euler_updates}
\end{equation}
where $z_0=\bar z_0$, $h>0$, and $k=0,\ldots,K-1$.

\begin{assumption}[Reference regularity and local alignment]
\label{ass:local_alignment}
There exist $L,\eta\geq0$ such that, at every sampling step,
\begin{align}
\|v_\star(x,t_k)-v_\star(y,t_k)\|_2
&\leq L\|x-y\|_2,
\label{eq:reference_lipschitz}\\
\|v_\phi(z_k,t_k)-v_\star(z_k,t_k)\|_2
&\leq\eta,
\label{eq:local_alignment_bound}
\end{align}
where $x,y\in\Omega$, and $\Omega$ is a state region containing
both trajectories.
\end{assumption}

\begin{proposition}[Propagation of local alignment error]
\label{thm:trajectory_deviation}
Under Assumption~\ref{ass:local_alignment}, the trajectories above
satisfy the following bound after $K$ updates:
\begin{equation}
\begin{aligned}
\|z_K-\bar z_K\|_2
&\leq C(L,T)\eta,\\
C(L,T)
&=
\begin{cases}
\dfrac{e^{LT}-1}{L}, & L>0,\\[5pt]
T, & L=0,
\end{cases}
\end{aligned}
\label{eq:trajectory_deviation}
\end{equation}
where $T=Kh$. The full proof is provided in
Appendix~\ref{app:trajectory_deviation}.
\end{proposition}

\section{Experiments}
\label{sec:experiments}

\subsection{Experimental Setup}

\begin{table*}[t]
\centering
\caption{
Quantitative comparison of unsafe concept erasure
on FLUX.1-dev across Nudity, Violence, and Blood.
ASR (\%) is evaluated on I2P, P4D, UnDiff, and RAB.
For P4D and UnDiff, adversarial prompts are adapted
to each method; thus, a single method-independent
FLUX.1-dev ASR is not applicable and is marked as ``--''.
N-FID measures neighboring benign concept preservation,
while COCO FID, CLIP Score, and SSIM assess
generation quality and fidelity, with the latter three
metrics averaged across the three concept categories.
}
\label{tab:flux_comparison}
\setlength{\tabcolsep}{2.3pt}
\renewcommand{\arraystretch}{1.13}
\scriptsize

\resizebox{\textwidth}{!}{
\begin{tabular}{@{}l*{18}{c}@{}}
\toprule

\multirow{2}{*}{\textbf{Method}}
& \multicolumn{5}{c}{\textbf{ Nudity (ASR,\%,$\downarrow$)}}
& \multicolumn{5}{c}{\textbf{Violence (ASR,\%,$\downarrow$)}}
& \multicolumn{5}{c}{\textbf{Blood (ASR,\%,$\downarrow$)}}
& \multicolumn{3}{c}{\textbf{General Preservation}} \\

\cmidrule(lr){2-6}
\cmidrule(lr){7-11}
\cmidrule(lr){12-16}
\cmidrule(l){17-19}

& I2P & P4D & UnDiff & RAB & N-FID $\downarrow$
& I2P & P4D & UnDiff & RAB & N-FID $\downarrow$
& I2P & P4D & UnDiff & RAB & N-FID $\downarrow$
& \makecell{COCO\\FID $\downarrow$}
& CS $\uparrow$
& SSIM $\uparrow$ \\

\midrule

FLUX.1-dev
& 15.7 & - & - & 89.7 & -
& 65.5 & - & - & 75.7 & -
& 57.3 & - & - & 83.5 & -
& - & 30.9 & 1.000 \\

EraseAnything
& \underline{7.5} & 40.4 & \underline{21.8}
& 29.4 & 24.7
& \textbf{33.2} & \underline{51.2}
& \underline{68.7} & \underline{51.5}
& 30.9
& 37.9 & 59.2 & \underline{52.8}
& 58.5 & 27.1
& \underline{21.8} & 30.2 & 0.584 \\

ESD
& 9.2 & 47.7 & 26.8 & \textbf{11.0}
& 27.6
& 54.9 & 63.8 & 80.9 & 66.8
& 35.2
& \underline{34.6} & 57.0 & 70.2 & 55.1
& 29.4
& 23.1 & 28.4 & 0.493 \\

MACE
& 8.5 & 41.2 & 28.5 & 32.1
& 25.1
& 50.7 & 59.0 & 69.8 & 57.4
& 31.8
& 38.8 & \underline{55.2} & 68.1 & 56.3
& 28.7
& 24.2 & 29.5 & 0.536 \\

SafeRoPE
& 7.7 & \underline{38.0} & 24.3 & 22.5
& 22.2
& 42.9 & 61.4 & 69.5 & 57.1
& 26.1
& 39.5 & 55.3 & 58.6 & \underline{45.9}
& 24.5
& 21.9 & \textbf{30.8} & \underline{0.672} \\

SAFREE
& 8.3 & 50.7 & 30.7 & 54.1
& 24.0
& 45.7 & 62.3 & 75.6 & 58.4
& 28.7
& 43.2 & 65.4 & 75.2 & 68.4
& 26.8
& 22.3 & \underline{30.6} & 0.612 \\

SLD
& 11.2 & 59.3 & 35.6 & 78.6
& 26.4
& 56.9 & 69.3 & 83.1 & 71.3
& 33.5
& 45.0 & 68.2 & 82.4 & 71.8
& 30.2
& 23.0 & 30.5 & 0.658 \\

CoGFD
& 8.1 & 42.8 & 25.3
& 31.6 & 23.9
& 42.6 & 57.8
& 72.4 & 55.2
& 28.4
& 37.2 & 59.1
& 62.3 & 53.6
& 26.1
& 22.7 & 30.0
& 0.664 \\

GrOCE 
& 8.1 & 41.5 & 26.2
& 29.0 & 23.1
& 40.2 & 56.1
& 69.9 & 52.4
& 27.5
& 34.4 & 57.2
& 60.7 & 51.3
& 25.4
& 22.3 & 30.1
& 0.715 \\

\midrule

\textbf{DGSR (Ours)}
& \textbf{3.2} & \textbf{31.1}
& \textbf{17.4} & \underline{19.8}
& 21.6
& \underline{34.8} & \textbf{47.2}
& \textbf{58.0} & \textbf{41.3}
& 23.8
& \textbf{23.5} & \textbf{38.5}
& \textbf{50.1} & \textbf{40.7}
& 21.7
& \textbf{21.5} & \textbf{30.8}
& \textbf{0.752} \\

\bottomrule
\end{tabular}
}
\end{table*}

\begin{figure*}
    \centering
    \includegraphics[width=0.8\linewidth]{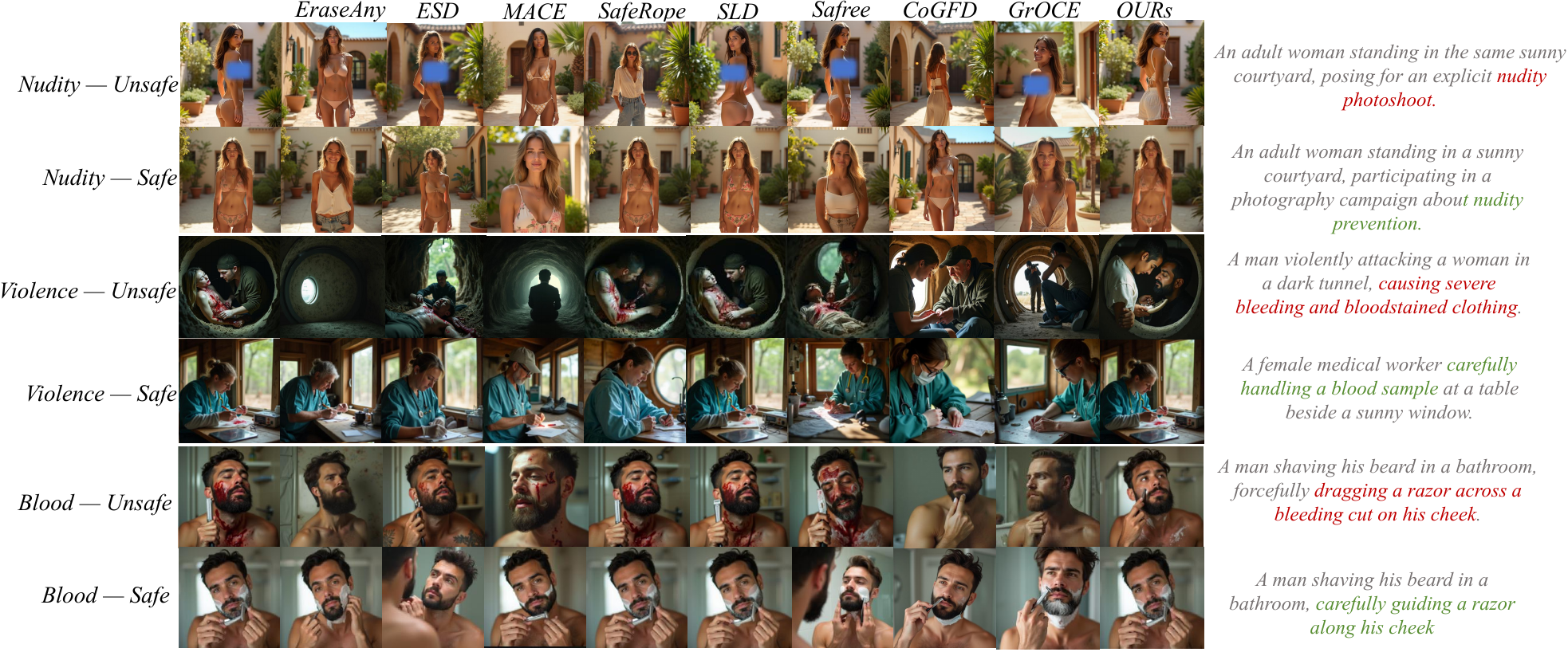}
    \caption{Qualitative comparison of unsafe and benign generation on FLUX.1-dev. We evaluate prompts involving nudity, violence, and blood. DGSR suppresses harmful semantic interactions while preserving benign visual content under related semantic conditions.}
    \label{fig:t2i_qualitative}
\end{figure*}

\begin{figure}[h]
    \centering
    \includegraphics[width=1.0\linewidth]{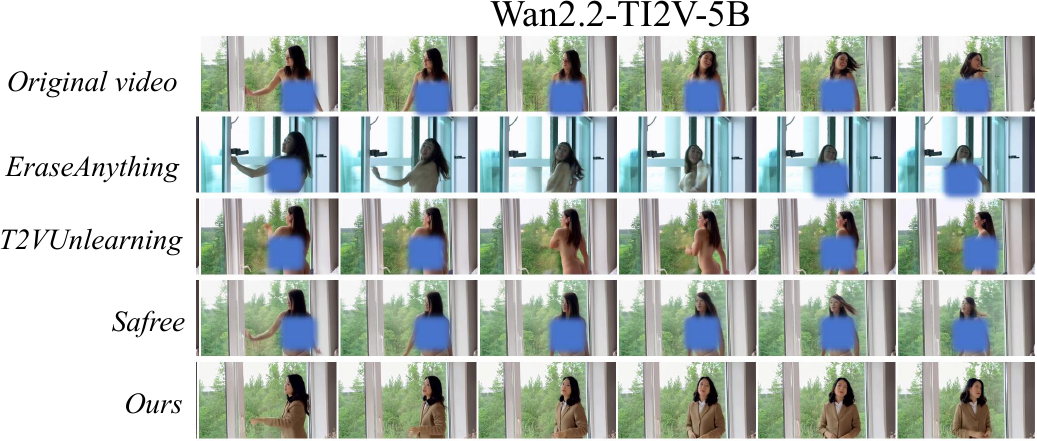}
     \caption{
Qualitative comparison of nudity on
Wan2.2-TI2V-5B.
}
    \label{fig:t2v_qualitative}
\end{figure}

\begin{figure}[h]
    \centering
    \includegraphics[width=1.0\linewidth]{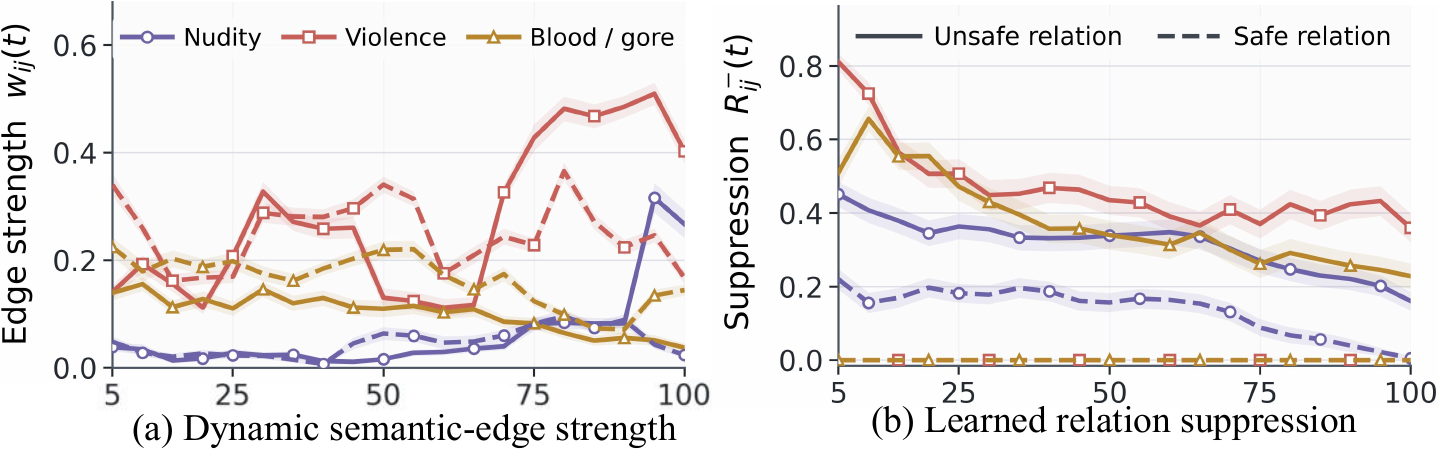}
    \caption{Visualization of dynamic semantic relations during sampling.(a) Evolution of semantic edge strengths across different unsafe categories. (b) Learned suppression strengths for unsafe (solid) and safe (dashed) relations.}
    \label{fig:dynamic_relation}
\end{figure}

\begin{table}[t]
\centering
\caption{
Quantitative comparison of unsafe concept erasure
on Open-Sora and Wan2.2-TI2V-5B.
We report nudity and blood/gore rates (\%)
on Gen and RAB,
along with subject consistency (Cons.).
}
\label{tab:video_comparison}
\setlength{\tabcolsep}{2.5pt}
\renewcommand{\arraystretch}{1.12}
\scriptsize

\begin{tabular*}{\columnwidth}{
@{\extracolsep{\fill}}lccccc@{}}
\toprule
\multirow{2}{*}{\textbf{Method}}
& \multicolumn{2}{c}{\textbf{Nudity $\downarrow$}}
& \multicolumn{2}{c}{\textbf{Blood/Gore $\downarrow$}}
& \multirow{2}{*}{
\shortstack{\textbf{Subj.}\\
\textbf{Cons. $\uparrow$}}} \\
\cmidrule(lr){2-3}
\cmidrule(lr){4-5}
& Gen & Ring-A-Bell & Gen & Ring-A-Bell & \\
\midrule

Open-Sora
& 58.29 & 19.21
& 61.38 & 35.75
& 95.92 \\

SAFREE
& 31.67 & 14.27
& 44.71 & 25.38
& 93.25 \\

EraseAnything++
& 17.10 & 7.71
& 27.63 & 14.21
& 94.18 \\

T2VUnlearning
& 16.50 & 9.53
& 30.83 & 12.67
& 93.20 \\

\textbf{DGSR (Ours)}
& \textbf{10.79} & \textbf{6.83}
& \textbf{24.92} & 13.04
& \textbf{95.04} \\

\midrule

Wan2.2-TI2V-5B
& 59.46 & 41.58
& 66.42 & 44.54
& 94.86 \\

SAFREE
& 52.67 & 36.00
& 61.21 & 39.08
& 92.11 \\

EraseAnything++
& 45.58 & 31.33
& 52.17 & 35.79
& 92.92 \\

T2VUnlearning
& 43.21 & 28.67
& 50.50 & 34.04
& 92.46 \\

\textbf{DGSR (Ours)}
& \textbf{20.92} & \textbf{12.04}
& \textbf{26.00} & \textbf{15.29}
& \textbf{94.13} \\

\bottomrule
\end{tabular*}
\end{table}

\begin{figure}[h]
    \centering
    \includegraphics[width=0.9\linewidth]{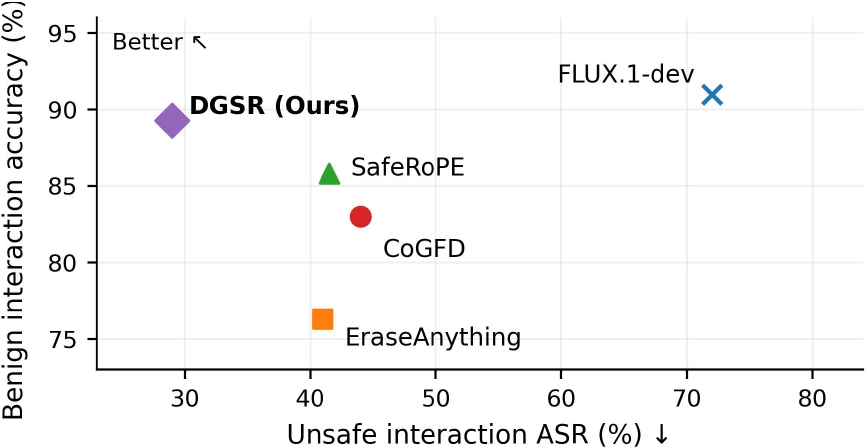}
    \caption{Counterfactual safety–preservation trade-off on FLUX. Unsafe interaction ASR and benign interaction accuracy are evaluated under the counterfactual protocol detailed in Appendix \ref{app:counterfactual_eval}.}
    \label{fig:tradeof}
\end{figure}










\textbf{Models and Baselines.}
We evaluate the safety suppression capability and semantic preservation performance of DGSR on both text-to-image (T2I) and text-to-video (T2V) generation tasks.
For T2I, we adopt FLUX.1-dev~\cite{flux2024} as the generation backbone and compare DGSR with ESD~\cite{gandikota2023erasing}, EraseAnything~\cite{gao2025eraseanything}, MACE~\cite{lu2024mace}, SafeRoPE~\cite{yang2026saferope},CoGFD~\cite{nie2025cogfd}, SAFREE~\cite{yoon2025safree}, GrOCE\cite{han2026groce}and SLD~\cite{schramowski2023safe}.
For T2V, we adopt Open-Sora~\cite{opensora} and Wan2.2-TI2V-5B~\cite{wan2025wan}, and compare DGSR with SAFREE~\cite{yoon2025safree}, EraseAnything++~\cite{gao2025eraseanything,fan2026eraseanything}, and T2VUnlearning~\cite{ye2025t2vunlearning}. Architecture, training details
are provided in Appendix~\ref{app:implementation}.

\textbf{Evaluation Benchmarks.}
For T2I, we evaluate three categories of unsafe content: Nudity, Violence, and Blood, using four benchmarks: I2P~\cite{schramowski2023safe}, Prompting4Debugging (P4D)~\cite{chin2024p4d},
UnlearnDiffAtk (UnDiff)~\cite{zhang2024unlearndiffatk}, and Ring-A-Bell (RAB)~\cite{tsai2024ringabell}, with 1,000 test prompts selected from each benchmark.
Additionally, we use MS-COCO-10K~\cite{lin2014coco} to evaluate general generation capabilities
and construct semantically neighboring concepts and paired benign prompts to assess non-target semantic preservation. For T2V, we evaluate Nudity and Gore on the Gen and RAB prompt sets, and assess temporal subject consistency on Open-Sora~\cite{opensora} and Wan2.2-TI2V-5B~\cite{wan2025wan}.


\textbf{Evaluation Metrics.}
For T2I, we employ the following metrics:
(1) Attack Success Rate (ASR) measures the proportion of unsafe generations;
(2) Neighbor FID (N-FID) evaluates the preservation of neighboring benign concepts;
(3) Fréchet Inception Distance (FID)~\cite{heusel2017fid} measures image generation quality;
(4) CLIP Score (CS)~\cite{hessel2021clipscore} evaluates text-image alignment;
(5) Structural Similarity Index Measure (SSIM)~\cite{wang2004ssim} measures structural similarity between paired benign generations and original model outputs.
Lower ASR, N-FID, and FID are preferred,
while higher CS and SSIM are better.
For T2V, we use
(1) Nudity/Gore Rate to measure the proportion of unsafe video frames, and
(2) Subject Consistency~\cite{huang2024vbench}
to evaluate subject coherence across frames.
NudeDetector~\cite{nudenet} and Q16~\cite{schramowski2022q16}
assess nudity and violence/blood/gore-related content,
while Qwen2.5-VL-7B-Instruct~\cite{bai2025qwen25vl}
evaluates harmful-interaction removal and shared-concept preservation.
Details are in Appendix~\ref{app:implementation}.

\subsection{Main Results}


\noindent\textbf{Text-to-Image Generation.}
As shown in Table~\ref{tab:flux_comparison},
DGSR achieves the lowest ASR in 10 of 12
evaluations across Nudity, Violence, and Blood,
while maintaining strong generation fidelity.
The qualitative results in Fig.~\ref{fig:t2i_qualitative}
 illustrate selective suppression of
harmful content and preservation of benign semantics.

\noindent\textbf{Text-to-Video Generation.}
DGSR reduces unsafe-frame rates on both video backbones
while maintaining high subject consistency
(Table~\ref{tab:video_comparison}).
Fig.~\ref{fig:t2v_qualitative} shows qualitative results;
Appendices \ref{app:dynamic_relation_analysis}--\ref{app:spatiotemporal_graph} provide video relation analyses,
and Appendix \ref{app:additional_qualitative} presents additional visual comparisons.

\noindent\textbf{Counterfactual Safety--Preservation Trade-off.}
As shown in Fig.~\ref{fig:tradeof}, we conduct a counterfactual evaluation using matched unsafe/benign prompt pairs that preserve shared concepts and scene context while modifying only the key semantic interaction. DGSR achieves a better safety--preservation trade-off, substantially reducing unsafe interaction ASR while retaining benign interaction accuracy close to the original model. A stricter Selective Erasure Success (SES) metric is further reported in Appendix~\ref{app:counterfactual_eval}.

\subsection{Analysis and Ablations}



\noindent\textbf{Component Ablations.}
Table~\ref{tab:ablation_violence} evaluates the contributions
of dynamic visual grounding, contextual graph reasoning,
and dual-level intervention.
Compared with full DGSR, Text-only Graph, Static Relation
Policy, and Pairwise MLP (w/o MP) consistently yield higher
ASR, demonstrating the benefits of visual-state-aware
relation modeling and contextual message passing.
Removing either semantic-level suppression or relation-level
recomposition degrades safety performance,
confirming their complementary contributions.
Comparisons of graph edge metrics are provided
in Appendix~\ref{app:edge_metric}.

\noindent\textbf{Relation Dynamics and Selective Suppression.}
Fig.~\ref{fig:dynamic_relation}(a) shows distinct
category-dependent variations in semantic edge strengths
throughout sampling, highlighting the evolving nature
of semantic interactions.
Fig.~\ref{fig:dynamic_relation}(b) shows that DGSR
assigns consistently higher suppression scores to unsafe
relations than to benign ones, demonstrating selective
relation-level intervention rather than uniform suppression.

\begin{table}[t]
\centering
\caption{
Ablation study of DGSR on FLUX.1-dev under the
Violence setting.
}
\label{tab:ablation_violence}

\setlength{\tabcolsep}{2.5pt}
\renewcommand{\arraystretch}{1.12}
\scriptsize

\resizebox{\columnwidth}{!}{%
\begin{tabular}{@{}lcccccc@{}}
\toprule
\multirow{2}{*}{\textbf{Variant}}
& \multicolumn{4}{c}{
\textbf{Violence ASR (\%) $\downarrow$}}
& \multicolumn{2}{c}{
\textbf{Preservation}} \\
\cmidrule(lr){2-5}
\cmidrule(l){6-7}
& I2P & P4D & UnDiff & RAB
& FID $\downarrow$ & CLIP $\uparrow$ \\
\midrule


\textbf{Full DGSR}
& \textbf{34.8} & \textbf{47.2}
& \textbf{58.0} & \textbf{41.3}
& 22.4 & 30.5 \\

\midrule
\multicolumn{7}{@{}l}{
\textit{Relation Modeling}} \\

Text-only Graph
& 52.4 & 64.7 & 72.6 & 59.1
& 23.1 & 30.2 \\

Static Relation Policy
& 57.6 & 62.9 & 63.4 & 56.8
& 22.2 & 30.3 \\

Pairwise MLP (w/o MP)
& 44.1 & 55.9 & 64.4 & 50.2
& 22.7 & 30.4 \\

\midrule
\multicolumn{7}{@{}l}{
\textit{Intervention Mechanism}} \\

w/o Semantic Suppression
& 43.7 & 55.6 & 61.5 & 44.9
& \textbf{21.8} & \textbf{30.7} \\

w/o Relation Suppression
& 39.5 & 57.0 & 65.1 & 52.3
& 23.8 & 30.1 \\

\bottomrule
\end{tabular}%
}
\end{table}

\begin{figure}[h]
    \centering
    \includegraphics[width=1.0\linewidth]{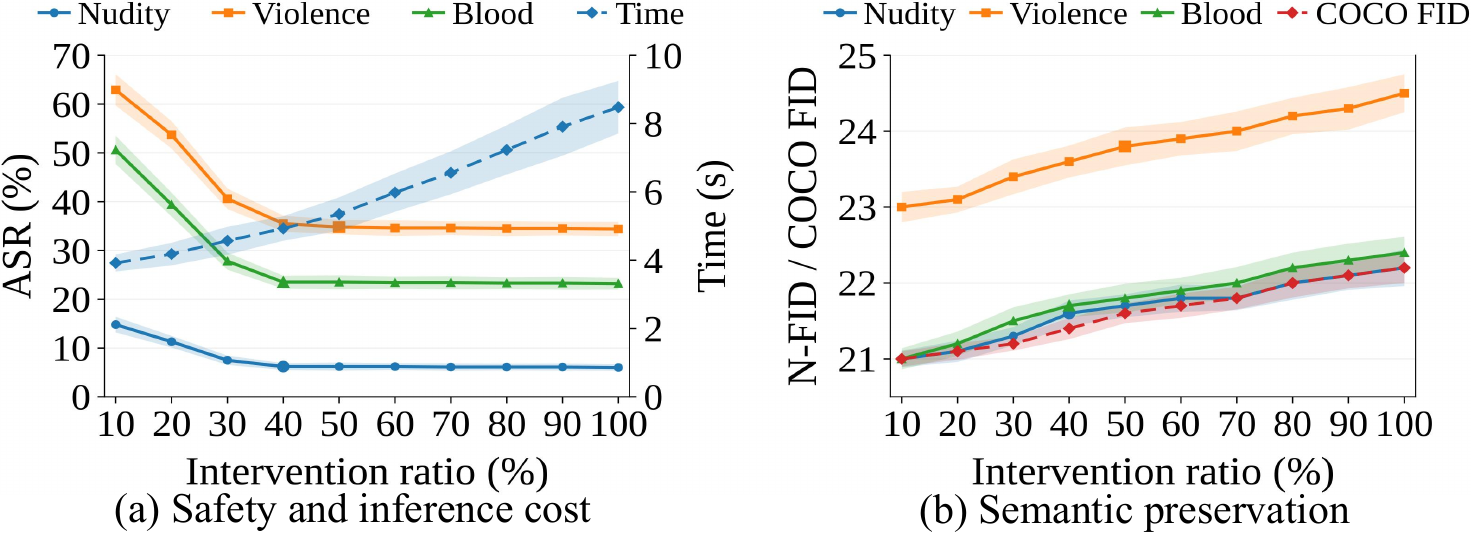}
    \caption{Effect of intervention ratio on safety,
semantic preservation, and inference efficiency.
(a) ASR together
with inference time under different intervention ratios.
(b) N-FID and COCO FID for general generation quality.}
    \label{fig:timecost}
\end{figure}

\noindent\textbf{Efficiency Trade-off.}
As shown in Fig.\ref{fig:timecost}, increasing the intervention ratio improves safety but gradually incurs higher inference cost and semantic degradation, with the gains eventually saturating.
We therefore adopt a moderate intervention ratio to balance safety, preservation, and efficiency; detailed computational-cost comparisons are provided in Appendix \ref{app:effeicent}.

\section{Conclusion}
\label{sec:conclusion}

We presented DGSR, which
addresses unsafe visual generation by modeling semantic interactions
as they evolve during generation. DGSR shifts safety control from
static suppression of isolated concepts to dynamic identification
and suppression of harmful semantic relations. By constructing
semantic graphs from current text--visual representations and
performing contextual relation reasoning, it updates relation-level
decisions throughout sampling and translates them into attention
interventions within a frozen backbone. Experiments on image and
video generation demonstrate effective unsafe-content suppression
while preserving generation quality and video subject consistency.
These findings highlight dynamic semantic-interaction modeling
as a promising approach to balancing generation safety and
semantic preservation.

\section{LLM Usage Statement}
ChatGPT (OpenAI) was used to assist with language
polishing, improving clarity, and refining academic
expression in the manuscript and supplementary material.
The authors reviewed the final text and take full
responsibility for its scientific content.

{
    \small
    \bibliographystyle{ieeenat_fullname}
    \bibliography{ref}
}

\clearpage

\appendix

\begingroup
\color{tocblue}  
\small
\setlength{\parindent}{0pt}
\setlength{\parskip}{2pt}

\newcommand{\suppTocSec}[2]{%
  \noindent\textbf{\ref{#1}\quad #2}%
  \nobreak\dotfill\pageref{#1}\par
}

\newcommand{\suppTocSub}[2]{%
  \noindent\hspace*{1em}%
  \ref{#1}\quad #2%
  \nobreak\dotfill\pageref{#1}\par
}

\noindent{\bfseries Supplementary Material Contents}
\par\smallskip

\suppTocSec{app:method}{Additional Method Details}
\suppTocSub{app:grounding}{Semantic Grounding and Supernode Features}
\suppTocSub{app:ot}{Wasserstein-based Edge Attributes}
\suppTocSub{app:gnn}{Edge-GNN Architecture}
\suppTocSub{app:counterfactual_supervision}{Counterfactual Supervision and Training Procedure}
\suppTocSub{app:semantic_reweighting}{Proof of Relative Semantic Reweighting}
\suppTocSub{app:trajectory_deviation}{Local Prediction Alignment and Sampling Deviation}

\medskip
\suppTocSec{app:implementation}{Implementation Details}
\suppTocSub{Architecture}{Model Architecture Details}
\suppTocSub{training_impl}{Training Implementation}
\suppTocSub{infer_impl}{Inference Implementation}
\suppTocSub{evaluation}{Safety Evaluation Protocol}

\medskip
\suppTocSec{app:additional_experiments}{Additional Experimental Analysis}
\suppTocSub{app:dynamic_relation_analysis}{Additional Dynamic Relation Analysis}
\suppTocSub{app:graph_risk_gate}{Graph Risk Gate Analysis}
\suppTocSub{app:spatiotemporal_graph}{Spatiotemporal Semantic Graph Construction}
\suppTocSub{app:counterfactual_eval}{Counterfactual Interaction Evaluation}
\suppTocSub{app:edge_metric}{Effect of Graph Edge Metrics}
\suppTocSub{app:effeicent}{Computational Efficiency}

\suppTocSub{app:additional_qualitative}{Additional Qualitative Results}

\endgroup

\medskip

\section{Additional Method Details}
\label{app:method}

\subsection{Semantic Grounding and Supernode Features}
\label{app:grounding}

\begin{figure}[h]
    \centering
    \includegraphics[width=0.85\linewidth]{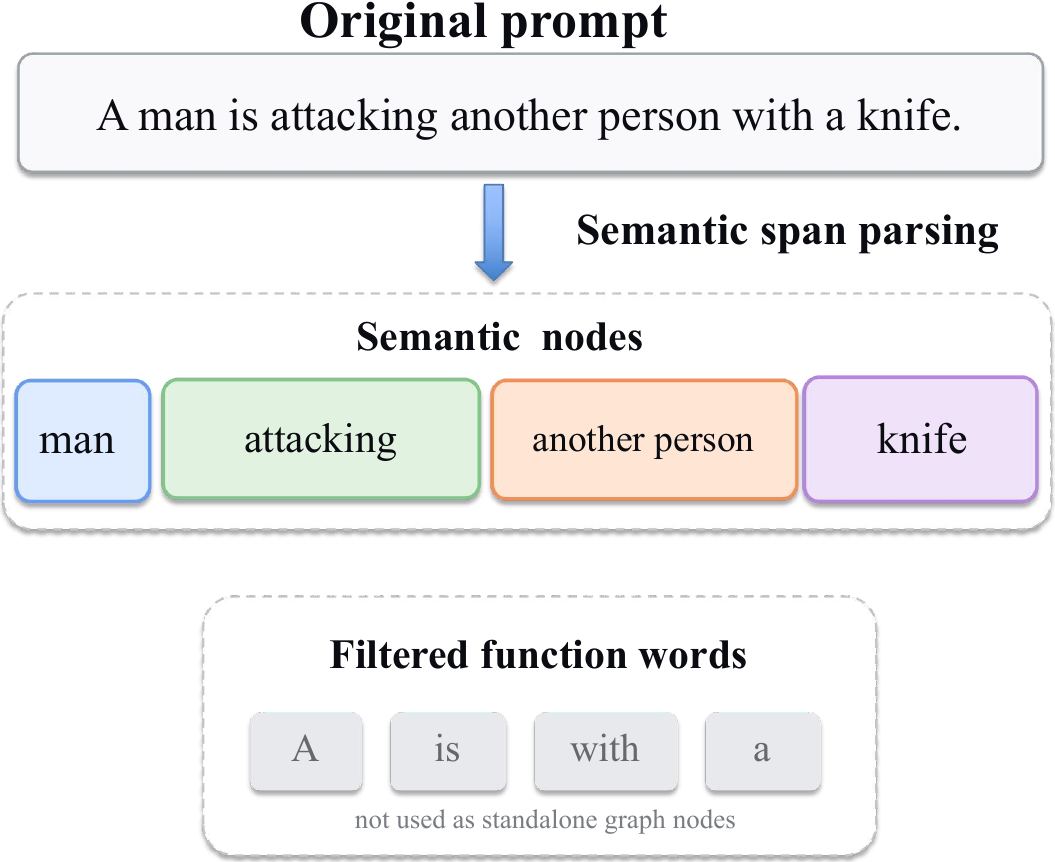}
\caption{
\textbf{Illustration of semantic span parsing in DGSR.}
The input prompt is segmented into meaningful semantic
spans, where related words (e.g., ``another person'')
are grouped into a single graph node.
Function words are excluded from standalone node
construction, while the original prompt remains
unchanged for text-conditioned generation.
}
\label{fig:semantic_parsing}
\end{figure}

\paragraph{Semantic Span Parsing and Token Alignment.}
As illustrated in Fig.~\ref{fig:semantic_parsing},
DGSR adopts NLP-based lexical segmentation and
stopword filtering to construct semantic-level
graph nodes rather than individual tokens.
Semantically coherent phrases, such as
``another person'', are grouped into single spans,
while function words are excluded from standalone nodes.
Each span is then aligned with its corresponding
T5 sub-tokens using tokenizer offset mappings.
Importantly, the original prompt remains unchanged,
ensuring that semantic filtering only affects
graph construction rather than text conditioning.

\paragraph{Semantic spans and cross-modal affinity.}
Given a sequence of text sub-tokens, we group the indices belonging to
semantic span $i$ into
$\mathcal S_i=\{k\mid k\text{ indexes a sub-token of span }i\}$,
where $i\in\{1,\ldots,N\}$ and $\mathcal S_i$ is nonempty.
For attention head $h$, we aggregate the text Keys within the span as
\begin{equation}
\bar{\mathbf K}_{i,h}
=\frac{1}{|\mathcal S_i|}\sum_{k\in\mathcal S_i}\mathbf K_{k,h},
\label{eq:app_span_key}
\end{equation}
where $\mathbf K_{k,h}\in\mathbb R^{D_h}$ is the Key of text token $k$.
For visual token $p$, its Query is
$\mathbf Q_{p,h}\in\mathbb R^{D_h}$.
Averaging over $H$ attention heads yields
\begin{equation}
Z_{p,i}=\frac{1}{H\sqrt{D_h}}
\sum_{h=1}^{H}\mathbf Q_{p,h}^{\top}\bar{\mathbf K}_{i,h}.
\end{equation}
We then normalize along the semantic-span and visual-position dimensions,
respectively:
\begin{equation}
\begin{aligned}
A_{p,i}&=\frac{\exp(Z_{p,i})}{\sum_{j=1}^{N}\exp(Z_{p,j})},\\
P_{p,i}&=\frac{A_{p,i}}{\sum_{q=1}^{M}A_{q,i}}.
\end{aligned}
\label{eq:app_normalization}
\end{equation}
where $A_{p,i}$ denotes the relative response of visual position $p$
to span $i$, while
$\mathbf P_i=(P_{1,i},\ldots,P_{M,i})^\top$
is the visual grounding distribution of that span, satisfying
$\sum_p P_{p,i}=1$.
Numerical safeguards may be applied to the normalization denominators
in practice, while distributions used for the standard Wasserstein
distance must remain normalized.

\paragraph{Supernode statistics.}
In addition to the mean-pooled textual features and grounding-weighted
visual features defined in the main paper, each node includes
response statistics:
\begin{equation}
\begin{aligned}
\mathcal H_i
&=-\frac{1}{\log M}\sum_{p=1}^{M}
P_{p,i}\log(P_{p,i}+\epsilon),\\
\mathbf x_i^{\mathrm{stat}}
&=\left[\max_p A_{p,i};\,
\frac{1}{M}\sum_{p=1}^{M}A_{p,i};\,
\mathcal H_i\right].
\end{aligned}
\label{eq:app_node_stats}
\end{equation}
where $\mathcal H_i$ is the normalized grounding entropy, and
$\epsilon>0$ prevents numerical issues arising from $\log 0$.
Concatenating the textual, visual, and statistical features gives
$\mathbf x_i\in\mathbb R^{d_{\mathrm{text}}+d_{\mathrm{visual}}+3}$,
where $d_{\mathrm{text}}$ and $d_{\mathrm{visual}}$ denote the
dimensions of the two hidden-state representations.
In our FLUX configuration, both dimensions are $3072$, yielding a
total node dimension of $6147$.

\paragraph{Span-level Value Construction.}
For relation-level semantic recomposition, we aggregate
the text-token Value representations within each semantic
span. Let $\mathbf V_k\in\mathbb R^{D_v}$ denote the
Value representation of text token $k$ in the current
attention module. The Value representation of span $i$
is computed as
\begin{equation}
\mathbf V_i^{\mathrm{span}}
=
\frac{1}{|\mathcal S_i|}
\sum_{k\in\mathcal S_i}\mathbf V_k.
\label{eq:app_span_value}
\end{equation}
Stacking the span representations gives
$\mathbf V_{\mathrm{span}}\in\mathbb R^{N\times D_v}$.
This matrix is used in Eq.~\eqref{eq:semantic_message}
to map relation-induced changes in semantic responses
to visual attention output increments.

\subsection{Spatial Wasserstein-based Edge Attributes}
\label{app:ot}

At sampling step $t$ and Transformer layer $l$, each
semantic node $i$ is associated with a normalized visual
grounding distribution $\mathbf P_i^{t,l}$, derived from
the current text--visual attention responses.
These distributions characterize where different
semantic units are grounded in the evolving visual
representation.

To quantify the spatial association between semantic
nodes, we employ the first-order Wasserstein distance
($W_1$). Unlike pointwise overlap measures, which
compare grounding responses at identical positions,
the Wasserstein distance additionally accounts for
the geometric displacement between response regions.
It therefore distinguishes spatially nearby
distributions from distant ones, even when their
pointwise overlap is similarly low.

\paragraph{Spatial Transport Cost.}
We first consider visual tokens arranged on an
$H\times W$ spatial grid, with $M=HW$.
Each visual position $p$ is assigned a normalized
spatial coordinate
\begin{equation}
\mathbf u_p =
\left(
\frac{x_p}{W-1},
\frac{y_p}{H-1}
\right).
\label{eq:app_spatial_coordinate}
\end{equation}
The transport cost between visual positions $p$
and $q$ is defined by their Euclidean distance:
\begin{equation}
C_{pq}=\|\mathbf u_p-\mathbf u_q\|_2.
\label{eq:app_transport_cost}
\end{equation}
This spatial cost assigns smaller penalties to
nearby positions and larger penalties to
spatially distant positions.

\paragraph{Wasserstein-based Association.}
Omitting $t,l$ for clarity, let
$\mathbf P_i,\mathbf P_j\in\mathbb R_{\ge 0}^{M}$
denote the normalized grounding distributions
of semantic nodes $i$ and $j$, satisfying
$\sum_p P_{p,i}=\sum_p P_{p,j}=1$.

Their spatial Wasserstein-1 distance is defined as
\begin{equation}
\begin{aligned}
d_{ij}
&=\min_{\boldsymbol\pi\in\Pi(\mathbf P_i,\mathbf P_j)}
\sum_{p=1}^{M}\sum_{q=1}^{M}
\pi_{pq}C_{pq},\\
\Pi(\mathbf P_i,\mathbf P_j)
&=\left\{
\boldsymbol\pi\ge0\;\middle|\;
\boldsymbol\pi\mathbf1=\mathbf P_i,\,
\boldsymbol\pi^\top\mathbf1=\mathbf P_j
\right\}.
\end{aligned}
\label{eq:app_ot}
\end{equation}
Here, $\boldsymbol\pi\in\mathbb R_{\ge0}^{M\times M}$
denotes a transport plan, where $\pi_{pq}$
represents the probability mass transported
from position $p$ to position $q$.
The constraints enforce the prescribed grounding
distributions as the two marginals.

The resulting distance measures the minimum
spatial transport cost required to align the
two grounding distributions.
A smaller $d_{ij}$ indicates stronger spatial
agreement, whereas a larger distance reflects
greater spatial separation.

Following Eq.~\eqref{eq:edge_weight} in the
main paper, we convert this distance into
an initial graph edge weight:
\begin{equation}
w_{ij}=\exp(-d_{ij}/\tau_{\mathrm{ot}}),
\label{eq:app_spatial_edge_weight}
\end{equation}
where $\tau_{\mathrm{ot}}>0$ controls the
distance sensitivity. For the normalized spatial transport cost defined in
Eq.~(32), we set $\tau_{\mathrm{ot}}=0.10$.
This scale parameter is kept fixed for the
reported comparisons.

Importantly, $w_{ij}$ represents geometric
grounding association rather than semantic
risk. The subsequent Edge-GNN combines these
association weights with semantic node features
and contextual information to predict directed
relation-level suppression strengths.
Although the initial geometric weights are
symmetric, the learned relation predictions
can be asymmetric due to the ordered endpoint
representations.

\paragraph{Extension to Video Grounding.}
For video generation, visual tokens span both
spatial and temporal positions. When the same
spatial transport cost is used, the distance
is equivalently defined over the temporal
marginals of the grounding distributions:
\begin{equation}
\bar P_{(x,y),i}
=
\sum_{f=1}^{F}P_{(f,x,y),i},
\label{eq:app_video_spatial_marginal}
\end{equation}
where $f$ indexes temporal token positions.

The spatial Wasserstein distance then measures
the geometric discrepancy between these
spatial grounding marginals without introducing
an explicit temporal transport penalty.
Temporal context is represented in the
spatiotemporal visual features, while the
semantic interaction graph is dynamically
updated along the joint video sampling trajectory.

\subsection{Edge-GNN Architecture}
\label{app:gnn}

Let $q\in\{0,1\}$ index the message-passing rounds.
We project the node feature $\mathbf x_i$ and edge weight $w_{ij}$
through $f_v$ and $f_e$, respectively, obtaining
$\mathbf h_i^{(0)}$ and $\mathbf g_{ij}^{(0)}$.
Each round first updates the edge states:
\begin{equation}
\mathbf g_{ij}^{(q+1)}=
\operatorname{LN}\!\left(
\mathbf g_{ij}^{(q)}+
\operatorname{MLP}_{e}^{(q)}
([\mathbf h_i^{(q)};\mathbf h_j^{(q)};\mathbf g_{ij}^{(q)}])
\right),
\label{eq:app_edge_update}
\end{equation}
where $\operatorname{MLP}_{e}^{(q)}$ is the edge update network and
$\operatorname{LN}$ denotes LayerNorm.
The updated edge states are then used to form messages:
\begin{equation}
\begin{aligned}
\mathbf m_{i\rightarrow j}^{(q)}
&=\operatorname{MLP}_{m}^{(q)}
([\mathbf h_i^{(q)};\mathbf g_{ij}^{(q+1)}]),\\
\bar{\mathbf m}_j^{(q)}
&=\sum_{i:(i,j)\in\mathcal E}
\mathbf m_{i\rightarrow j}^{(q)},
\end{aligned}
\label{eq:app_message}
\end{equation}
where $\mathbf m_{i\rightarrow j}^{(q)}$ is the relation-conditioned
message passed from node $i$ to node $j$, and
$\bar{\mathbf m}_j^{(q)}$ is the aggregated message at node $j$.
The node states are subsequently updated as
\begin{equation}
\mathbf h_j^{(q+1)}=
\operatorname{LN}\!\left(
\mathbf h_j^{(q)}+
\operatorname{MLP}_{v}^{(q)}
([\mathbf h_j^{(q)};\bar{\mathbf m}_j^{(q)}])
\right),
\label{eq:app_node_update}
\end{equation}
where $\operatorname{MLP}_{m}^{(q)}$ and
$\operatorname{MLP}_{v}^{(q)}$ are the message and node update
networks, respectively.
After two rounds, the resulting node and edge states are
$\mathbf h_i^{(2)}$ and $\mathbf g_{ij}^{(2)}$.
The relation suppression head in Eq.~\eqref{eq:relation_score}
then predicts $R_{ij}^-\in[0,1]$.
The auxiliary graph-level classification head operates on the
Edge-GNN graph representation and is used only in the first training
stage, whereas the relation suppression head is used in subsequent
training stages and at inference time.

\subsection{Counterfactual Supervision and Training Procedure}
\label{app:counterfactual_supervision}

\noindent\textbf{Counterfactual Pair Construction.}
For each unsafe text condition $c_u$, we construct a corresponding
safe condition $c_s$ by modifying the key harmful action or
interaction target while preserving the subjects, objects, and scene
context as much as possible.
Candidate relations are identified from the semantic interactions
modified between the paired texts, rather than matched by token
positions or visual association strengths.
Semantic spans are parsed separately for the two conditions, so the
paired prompts need not have identical token counts or graph sizes.

\noindent\textbf{Focused Relation Selection.}
For each text pair, we identify the candidate relation sets
$\mathcal{E}_u^f$ and $\mathcal{E}_s^f$ from the target actions and
their participating entities.
For example, in the Violence setting,
``stab the woman with a knife'' and
``slice the bread with a knife'' yield the following focused relations:
\[
\begin{aligned}
\mathcal{E}_u^f
=&\{(\mathrm{knife},\mathrm{stab}),
(\mathrm{stab},\mathrm{woman})\},\\
\mathcal{E}_s^f
=&\{(\mathrm{knife},\mathrm{slice}),
(\mathrm{slice},\mathrm{bread})\}.
\end{aligned}
\]
Relations unrelated to the modified target interaction remain in
the dynamic graph for contextual reasoning but are excluded from
the focused relation contrastive supervision for the current pair.
Samples for which the target interaction cannot be reliably identified,
or for which the candidate relation set is empty, are excluded from
this training stage.

\noindent\textbf{Training Configuration.}
The paired conditions share the latent state $\mathbf{x}_t$,
sampling step $t$, and Transformer layer $l$.
Relation risk scores are computed by a prediction network with
shared parameters.
In Eq.~\eqref{eq:relation_loss}, we empirically set the contrastive
margin to $m=0.3$ to encourage sufficient separation between the
suppression scores of unsafe and benign interactions.
This value is treated as an empirical hyperparameter and does not
assume any particular ordering of visual association strengths
between the two conditions.
An example of the paired texts, candidate relations, and contrastive
supervision computation is provided in
Table~\ref{tab:pseudo_label_example}.

\subsection{Proof of Relative Semantic Reweighting}
\label{app:semantic_reweighting}

\begin{proof}[Proof of Proposition~\ref{thm:semantic_reweighting}]
Fix the sampling state, Transformer layer, and visual position $p$,
and let $N$ denote the number of semantic spans.
By the definition of the relation branch, the original semantic
response is
\begin{equation}
A_{p,i}
=
\frac{\exp(Z_{p,i})}
{\sum_{m=1}^{N}\exp(Z_{p,m})}.
\label{eq:app_original_response}
\end{equation}
The relation-induced bias and the adjusted response are defined as
\begin{equation}
\begin{aligned}
S_{p,i}
&=
\sum_{m=1}^{N}A_{p,m}R^-_{im},\\
\widetilde A_{p,i}
&=
\frac{\exp(Z_{p,i}-S_{p,i})}
{\sum_{m=1}^{N}\exp(Z_{p,m}-S_{p,m})}.
\end{aligned}
\label{eq:app_adjusted_response}
\end{equation}

Let the normalization factor of the original softmax be
\[
C_p=\sum_{m=1}^{N}\exp(Z_{p,m}).
\]
From Eq.~\eqref{eq:app_original_response}, we have
$\exp(Z_{p,i})=C_p A_{p,i}$.
Substituting this identity into Eq.~\eqref{eq:app_adjusted_response}
yields
\begin{equation}
\begin{aligned}
\widetilde A_{p,i}
&=
\frac{\exp(Z_{p,i})\exp(-S_{p,i})}
{\sum_{m=1}^{N}
 \exp(Z_{p,m})\exp(-S_{p,m})}\\
&=
\frac{C_p A_{p,i}\exp(-S_{p,i})}
{\sum_{m=1}^{N}
 C_p A_{p,m}\exp(-S_{p,m})}\\
&=
\frac{A_{p,i}\exp(-S_{p,i})}
{\sum_{m=1}^{N}A_{p,m}\exp(-S_{p,m})}.
\end{aligned}
\label{eq:app_response_reweighting}
\end{equation}

The finite logits in Assumption~\ref{ass:finite_logits}, together
with the bounded relation scores defined in the main paper,
ensure that all components of both the original and adjusted
softmax responses are strictly positive.
Thus, for any two semantic spans $i,j$, we may take the ratio of
the corresponding components in Eq.~\eqref{eq:app_response_reweighting}.
Since they share the same normalization denominator,
\begin{equation}
\begin{aligned}
\frac{\widetilde A_{p,i}}{\widetilde A_{p,j}}
&=
\frac{A_{p,i}\exp(-S_{p,i})}
{A_{p,j}\exp(-S_{p,j})}\\
&=
\frac{A_{p,i}}{A_{p,j}}
\exp\!\left[-(S_{p,i}-S_{p,j})\right].
\end{aligned}
\label{eq:app_relative_response_ratio}
\end{equation}

When $S_{p,i}>S_{p,j}$, the exponential factor is smaller than $1$,
so the response ratio of span $i$ relative to span $j$ decreases.
When $S_{p,i}=S_{p,j}$, the exponential factor equals $1$,
and the ratio remains unchanged.
This completes the proof of Proposition~\ref{thm:semantic_reweighting}.
\end{proof}

\paragraph{Meaning of the relation direction.}
The relation mapping $S=A(R^-)^\top$ can be written elementwise as
\begin{equation}
\begin{aligned}
S_{p,i}
&=
\sum_{j=1}^{N}
A_{p,j}\big[(R^-)^\top\big]_{j,i}\\
&=
\sum_{j=1}^{N}A_{p,j}R^-_{ij}.
\end{aligned}
\label{eq:app_relation_direction}
\end{equation}
Thus, $R^-_{ij}$ contributes to the penalty assigned to source span
$i$ at visual position $p$, weighted by the response $A_{p,j}$ of
context span $j$ at that position.
With other quantities fixed, the same relation score makes a larger
contribution to the penalty at positions with larger $A_{p,j}$.
The change in the response of span $i$ relative to other spans is
then determined by the penalty difference in
Eq.~\eqref{eq:app_relative_response_ratio}.

These results characterize the relative reweighting of semantic
responses induced by the relation branch.
Because the softmax renormalizes the responses, they do not imply
that every response component decreases.
In particular, if all spans at a visual position receive the same
penalty, the adjusted semantic response is identical to the original
response.
The accuracy of the predicted relation scores and the semantic
preservation achieved by the complete two-level intervention are
further evaluated experimentally.

\subsection{Local Prediction Alignment and Sampling Deviation}
\label{app:trajectory_deviation}

\paragraph{Sampling setup.}
We use a time parameter that increases along the sampling direction.
If the original model adopts a decreasing noise-time convention,
we absorb the sign induced by the change of time direction into
the definition of the predicted field.
Let $z_k$ and $\bar z_k$ denote the latent states of the intervened
and reference trajectories, respectively, at step $k$,
distinguishing them from the noise-time notation $x_t$ used in
the main paper.

Consider fixed-step Euler updates with the same initial state,
the same time grid $t_k=t_0+kh$, and step size $h>0$:
\begin{equation}
\begin{aligned}
z_{k+1}
&=
z_k+h\,v_\phi(z_k,t_k),\\
\bar z_{k+1}
&=
\bar z_k+h\,v_\star(\bar z_k,t_k),\\
z_0&=\bar z_0,
\end{aligned}
\label{eq:app_euler_updates}
\end{equation}
where $v_\phi$ is the intervened field, $v_\star$ is the reference
field, $k=0,\ldots,K-1$, and the total sampling interval has
length $T=Kh$.

By Assumption~\ref{ass:local_alignment}, there exist a state region
$\Omega$ containing both trajectories and a constant $L\geq0$ such
that
\begin{equation}
\|v_\star(x,t_k)-v_\star(y,t_k)\|_2
\leq L\|x-y\|_2,
\quad x,y\in\Omega.
\label{eq:app_reference_lipschitz}
\end{equation}
Moreover, at every state along the intervened trajectory,
\begin{equation}
\|v_\phi(z_k,t_k)-v_\star(z_k,t_k)\|_2
\leq\eta,
\label{eq:app_local_prediction_error}
\end{equation}
where $\eta\geq0$.
Importantly, Eq.~\eqref{eq:app_local_prediction_error} compares the
outputs of the two fields at the same state $z_k$, rather than
comparing their outputs at states from different trajectories.

\begin{proof}[Proof of Proposition~\ref{thm:trajectory_deviation}]
Define the trajectory deviation at step $k$ as
\[
e_k=\|z_k-\bar z_k\|_2.
\]
Because the initial states are identical, $e_0=0$.

Subtracting the two updates in Eq.~\eqref{eq:app_euler_updates}
and adding and subtracting the reference prediction
$v_\star(z_k,t_k)$ at the same state gives
\begin{equation}
\begin{aligned}
z_{k+1}-\bar z_{k+1}
={}&z_k-\bar z_k\\
&+h\big[
v_\phi(z_k,t_k)-v_\star(z_k,t_k)
\big]\\
&+h\big[
v_\star(z_k,t_k)-v_\star(\bar z_k,t_k)
\big].
\end{aligned}
\label{eq:app_error_decomposition}
\end{equation}
The first prediction difference represents the local alignment
error at the same state, whereas the second captures the change
in the reference field caused by different input states.

Applying the triangle inequality to
Eq.~\eqref{eq:app_error_decomposition}, followed by
Eqs.~\eqref{eq:app_reference_lipschitz} and
\eqref{eq:app_local_prediction_error}, yields
\begin{equation}
\begin{aligned}
e_{k+1}
\leq{}&
e_k
+h\|v_\phi(z_k,t_k)-v_\star(z_k,t_k)\|_2\\
&+h\|v_\star(z_k,t_k)
      -v_\star(\bar z_k,t_k)\|_2\\
\leq{}&
e_k+h\eta+hLe_k\\
={}&
(1+hL)e_k+h\eta.
\end{aligned}
\label{eq:app_error_recurrence}
\end{equation}

Starting from $e_0=0$, the local error bound $h\eta$ introduced
at each step propagates through subsequent updates.
After $K$ steps,
\begin{equation}
\begin{aligned}
e_K
&\leq
(1+hL)^K e_0
+h\eta\sum_{k=0}^{K-1}(1+hL)^{K-1-k}\\
&=
h\eta\sum_{r=0}^{K-1}(1+hL)^r.
\end{aligned}
\label{eq:app_accumulated_error}
\end{equation}

For $L>0$, the geometric series satisfies
\[
\sum_{r=0}^{K-1}(1+hL)^r
=
\frac{(1+hL)^K-1}{hL}.
\]
Therefore,
\begin{equation}
\begin{aligned}
e_K
&\leq
h\eta\frac{(1+hL)^K-1}{hL}\\
&=
\frac{\eta}{L}
\big[(1+hL)^K-1\big].
\end{aligned}
\label{eq:app_discrete_error_bound}
\end{equation}
Using $1+hL\leq e^{hL}$ and $T=Kh$, we further obtain
\begin{equation}
e_K
\leq
\frac{\eta}{L}\big(e^{LKh}-1\big)
=
\frac{\eta}{L}\big(e^{LT}-1\big).
\label{eq:app_positive_lipschitz_bound}
\end{equation}

For $L=0$, Eq.~\eqref{eq:app_error_recurrence} reduces to
\[
e_{k+1}\leq e_k+h\eta.
\]
Summing from $e_0=0$ gives
\begin{equation}
e_K\leq Kh\eta=T\eta.
\label{eq:app_zero_lipschitz_bound}
\end{equation}

Combining both cases yields
\begin{equation}
\begin{aligned}
\|z_K-\bar z_K\|_2
&\leq C(L,T)\eta,\\
C(L,T)
&=
\begin{cases}
\dfrac{e^{LT}-1}{L}, & L>0,\\[5pt]
T, & L=0.
\end{cases}
\end{aligned}
\label{eq:app_final_trajectory_bound}
\end{equation}
This completes the proof of Proposition~\ref{thm:trajectory_deviation}.
\end{proof}

\paragraph{Connection to generation alignment.}
For paired unsafe and safe conditions $c_u$ and $c_s$, respectively,
at the same latent state $x_t$ and time $t$, we abbreviate the three
predictions as
\begin{equation}
\begin{aligned}
y_u
&=F_\theta(x_t,t,c_u),\\
y_s
&=F_\theta(x_t,t,c_s),\\
y_\phi
&=F_{\theta,\phi}^{\mathrm{DGSR}}(x_t,t,c_u).
\end{aligned}
\label{eq:app_prediction_definitions}
\end{equation}
By the increment definitions in the main paper,
\begin{equation}
\Delta=y_\phi-y_u,
\qquad
\Delta^*=y_s-y_u.
\label{eq:app_increment_definitions}
\end{equation}
Thus, the common unsafe-condition reference prediction $y_u$
cancels in the difference between increments:
\begin{equation}
\begin{aligned}
\Delta-\Delta^*
&=(y_\phi-y_u)-(y_s-y_u)\\
&=y_\phi-y_s.
\end{aligned}
\label{eq:app_increment_cancellation}
\end{equation}

Flattening the output tensors and denoting the number of elements
by $D$, the squared-error term satisfies
\begin{equation}
\begin{aligned}
\operatorname{MSE}(\Delta,\Delta^*)
&=
\frac{1}{D}
\|\operatorname{vec}(\Delta-\Delta^*)\|_2^2\\
&=
\frac{1}{D}
\|\operatorname{vec}(y_\phi-y_s)\|_2^2.
\end{aligned}
\label{eq:app_alignment_mse_equivalence}
\end{equation}
Hence, the squared-error component of increment matching is
equivalent to matching the intervened prediction to the
safe-condition reference prediction at the same state.
The sample-dependent normalization in the main-text loss remains
in effect, while the cosine term additionally constrains the
direction of the actual change relative to the target change.
This equivalence does not replace the full alignment loss with an
unweighted prediction MSE.

Under the unified sampling-time convention, the corresponding
intervened and reference fields are
\begin{equation}
\begin{aligned}
v_\phi(x,t)
&=
F_{\theta,\phi}^{\mathrm{DGSR}}(x,t,c_u),\\
v_\star(x,t)
&=
F_\theta(x,t,c_s).
\end{aligned}
\label{eq:app_alignment_fields}
\end{equation}
By Eq.~\eqref{eq:app_increment_cancellation}, the increment error
used in training corresponds to the difference between these two
fields at a shared state.
Proposition~\ref{thm:trajectory_deviation} further shows that if
this local discrepancy is uniformly controlled at the states
visited during sampling, and the reference field satisfies the
corresponding Lipschitz condition, then the accumulated deviation
between their trajectories is also bounded.

\paragraph{Connection to benign preservation.}
For a benign text condition $c_b$, let
\begin{equation}
\begin{aligned}
v_\phi(x,t)
&=
F_{\theta,\phi}^{\mathrm{DGSR}}(x,t,c_b),\\
v_\star(x,t)
&=
F_\theta(x,t,c_b).
\end{aligned}
\label{eq:app_preservation_fields}
\end{equation}
The benign preservation loss constrains the deviation of the
intervened prediction from the original model prediction under
the same benign condition and latent state.
If Assumption~\ref{ass:local_alignment} also holds for this pair
of prediction fields, the same error recurrence and trajectory
deviation bound apply.
This interpretation leaves the generation alignment and benign
preservation objectives in the main paper unchanged.

\paragraph{Scope of the analysis.}
The trajectory bound applies to a fixed-step Euler discretization
with identical initial states and a shared time grid.
Other samplers require an error recurrence derived from their own
update rules; the coefficient obtained here cannot be applied
directly.
The local error condition must hold at states actually visited
during sampling, and a small average training loss does not
by itself guarantee this pointwise condition.
Moreover, the bound controls latent-state distance and does not
directly guarantee perceptual similarity or the safety of the
final output.
The practical benefits of dynamic graph modeling and the accuracy
of predicted relation scores are therefore evaluated experimentally.

\begin{table*}[t]
\caption{
\textbf{Illustrative violence pair for
counterfactual relation contrastive supervision.}
The example demonstrates the construction
of paired unsafe/safe prompts, semantic interaction
correspondence, and focused relation sets.
The contrastive supervision distinguishes
harmful and benign interactions without
requiring one-to-one token alignment.
}

\centering
\small
\renewcommand{\arraystretch}{1.2}

\begin{minipage}{0.96\textwidth}
\raggedright

\hrule
\medskip

\textbf{1. Paired Texts}

\smallskip
\textbf{Example ID:}
\texttt{violence\_pair\_01}

\par\smallskip
\textbf{Unsafe prompt:}
In a kitchen, an adult man and an adult woman
stand beside a counter with a loaf of bread.
The man uses a knife to \textbf{stab the woman}.

\par\smallskip
\textbf{Safe counterpart:}
In a kitchen, an adult man and an adult woman
stand beside a counter with a loaf of bread.
The man uses a knife to \textbf{slice the bread}.

\par\smallskip
\textbf{Relation edit:}
knife--stab--woman
$\rightarrow$
knife--slice--bread.

\par\medskip
\hrule
\medskip

\textbf{2. Semantic Interaction Correspondence}

\smallskip
Both prompts preserve the semantic entities
\emph{man}, \emph{woman}, \emph{knife},
and \emph{bread}.
The unsafe prompt describes an interpersonal
violent action, whereas the safe counterpart
describes a benign food-preparation action.

The correspondence is defined by the modified
interaction roles rather than one-to-one token
or node alignment.

\par\medskip
\hrule
\medskip

\textbf{3. Counterfactual Relation Sets}

\smallskip
Based on the modified action and its participants,
the corresponding focus relation sets are

\[
\begin{aligned}
\mathcal{E}_u^f
=&\{
(\mathrm{knife},\mathrm{stab}),
(\mathrm{stab},\mathrm{woman})
\},\\
\mathcal{E}_s^f
=&\{
(\mathrm{knife},\mathrm{slice}),
(\mathrm{slice},\mathrm{bread})
\}.
\end{aligned}
\]

The unsafe relation set captures the
interaction between the weapon, violent action,
and target person.
In contrast, the safe relation set represents
the benign interaction between the tool,
food-preparation action, and target object.

Shared contextual relations, such as
man--knife, remain in the semantic graph
for contextual reasoning but are excluded
from the focused contrastive supervision.

\par\medskip
\hrule

\end{minipage}

\label{tab:pseudo_label_example}

\end{table*}

\begin{algorithm}[t]
\caption{Progressive Training and Inference of DGSR}
\label{alg:dgsr_reorganized}
\begin{algorithmic}[1]

\REQUIRE Frozen backbone $F_\theta$;
labeled data $\mathcal D_{\mathrm{disc}}$;
unsafe/safe pairs $\mathcal D_{\mathrm{pair}}$;
benign data $\mathcal D_{\mathrm{benign}}$;
development data $\mathcal D_{\mathrm{dev}}$

\ENSURE Trained intervention module $\phi$

\STATE \textbf{Stage I: Graph-level risk learning}

\FOR{labeled sample $(c,b)\in\mathcal D_{\mathrm{disc}}$}
    \STATE Extract $\mathcal G_t^l$ from frozen $F_\theta$
    \STATE Update graph encoder using
    $\mathcal L_{\mathrm{graph}}$
\ENDFOR

\STATE \textbf{Stage II: Counterfactual relation supervision}

\FOR{$(c_u,c_s)\in\mathcal D_{\mathrm{pair}}$}

    \STATE Construct $\mathcal G_u,\mathcal G_s$
    under shared $\mathbf x_t,t,l$

    \STATE Identify corresponding interaction sets
    $\mathcal E_u^f,\mathcal E_s^f$

    \STATE Predict relation scores
    $\mathbf R_u^-,\mathbf R_s^-$

    \STATE Aggregate interaction risks
    $s_u,s_s$ over $\mathcal E_u^f,\mathcal E_s^f$

    \STATE Update $\phi$ using
    $\mathcal L_{\mathrm{relation}}$

\ENDFOR

\STATE \textbf{Stage III: Generation alignment and preservation}

\FOR{paired unsafe/safe and benign conditions}

    \STATE Compute $\Delta^*,\Delta$
    and $\mathcal L_{\mathrm{retain}}$

    \STATE Update $\phi$ using
    $\mathcal L_{\mathrm{III}}$

\ENDFOR

\STATE Select inference settings on
$\mathcal D_{\mathrm{dev}}$

\STATE \textbf{Inference}

\STATE \textbf{Input:} Prompt $c$ and initial noise $\mathbf x_T$

\FOR{sampling step $t=T,\ldots,1$}

    \FOR{selected Transformer block $l$}

        \STATE Construct $\mathcal G_t^l$
        and predict $\mathbf R^-$

        \STATE Compute
        $s_G\gets
        \operatorname{Mean}
        (\operatorname{TopK}(\mathbf R^-,k_g))$

        \IF{$s_G\ge\tau_{\mathrm{gate}}$}

            \STATE Apply semantic-level
            token suppression

            \STATE Apply relation-level
            semantic message recomposition

        \ENDIF

    \ENDFOR

    \STATE Update latent using
    the intervened backbone

\ENDFOR

\STATE \textbf{Output:} Generated sample $\mathbf x_0$

\end{algorithmic}
\end{algorithm}

\section{Implementation Details}
\label{app:implementation}

\subsection{Network Architecture.}
\label{Architecture}
As summarized in Table~\ref{tab:dgsr_architecture},
DGSR encodes semantic spans into multimodal supernodes
and constructs pairwise relation representations.
A two-layer Edge-GNN performs contextual reasoning
through edge updates, edge-conditioned message passing,
and node updates.
The learned graph representations support graph-level
risk classification and relation-level suppression
prediction, while the pretrained backbone remains frozen.

\begin{table}[h]
\centering
\caption{
Architecture of the DGSR auxiliary network.
}
\label{tab:dgsr_architecture}
\setlength{\tabcolsep}{4pt}
\renewcommand{\arraystretch}{1.12}
\footnotesize

\begin{tabularx}{\columnwidth}{@{}lXr@{}}
\toprule
\textbf{Component} &
\textbf{Architecture / Operation} &
\textbf{Size} \\
\midrule

\multicolumn{3}{@{}l}{
\textit{Semantic Supernode Encoding}} \\

Text features
& Mean pooling of T5 span tokens
& 3072 \\

Visual features
& Grounding-weighted visual pooling
& 3072 \\

Response statistics
& Maximum, mean, entropy
& 3 \\

Node representation
& Feature concatenation
& 6147 \\

\addlinespace[4pt]
\multicolumn{3}{@{}l}{
\textit{Semantic Relation Encoding}} \\

Candidate edges
& Fully connected directed graph
& $N(N-1)$ \\

Edge attributes
& Wasserstein grounding similarity
& 1 \\

Edge encoder
& Learnable projection
& 64 \\

\addlinespace[4pt]
\multicolumn{3}{@{}l}{
\textit{Contextual Graph Reasoning}} \\

Edge update
& Residual MLP + LayerNorm
& \multirow{3}{*}{2 layers} \\

Message passing
& Edge-conditioned aggregation
& \\

Node update
& Residual MLP + LayerNorm
& \\

\addlinespace[4pt]
\multicolumn{3}{@{}l}{
\textit{Risk Prediction}} \\

Graph classifier
& Graph-level binary prediction
& 1 \\

Suppression head
& Directed edge-wise MLP
& $R^{-}$ \\

\bottomrule
\end{tabularx}
\end{table}

\subsection{Training Implementation.}
\label{training_impl}
We adopt a three-stage training strategy with the AdamW optimizer.
For graph-level risk learning, we use a learning rate of
$1\times10^{-4}$, a batch size of 16, and 10,000 training steps.
For relation-level supervision, the learning rate is set to
$5\times10^{-5}$ with a batch size of 8 for 8,000 steps.
For online generation alignment, we use a learning rate of
$1\times10^{-5}$, a batch size of 1, and 2,000 training steps.
The loss weights
$\lambda_{\mathrm{graph}}$,
$\lambda_{\mathrm{relation}}$,
$\lambda_{\mathrm{align}}$, and
$\lambda_{\mathrm{retain}}$
are all initialized to 1.0.
The pretrained FLUX backbone remains frozen
throughout training. For reproducibility, we use random seed $42$
for the three training stages.

\begin{figure*}[t]
    \centering
    \includegraphics[width=0.85\linewidth]{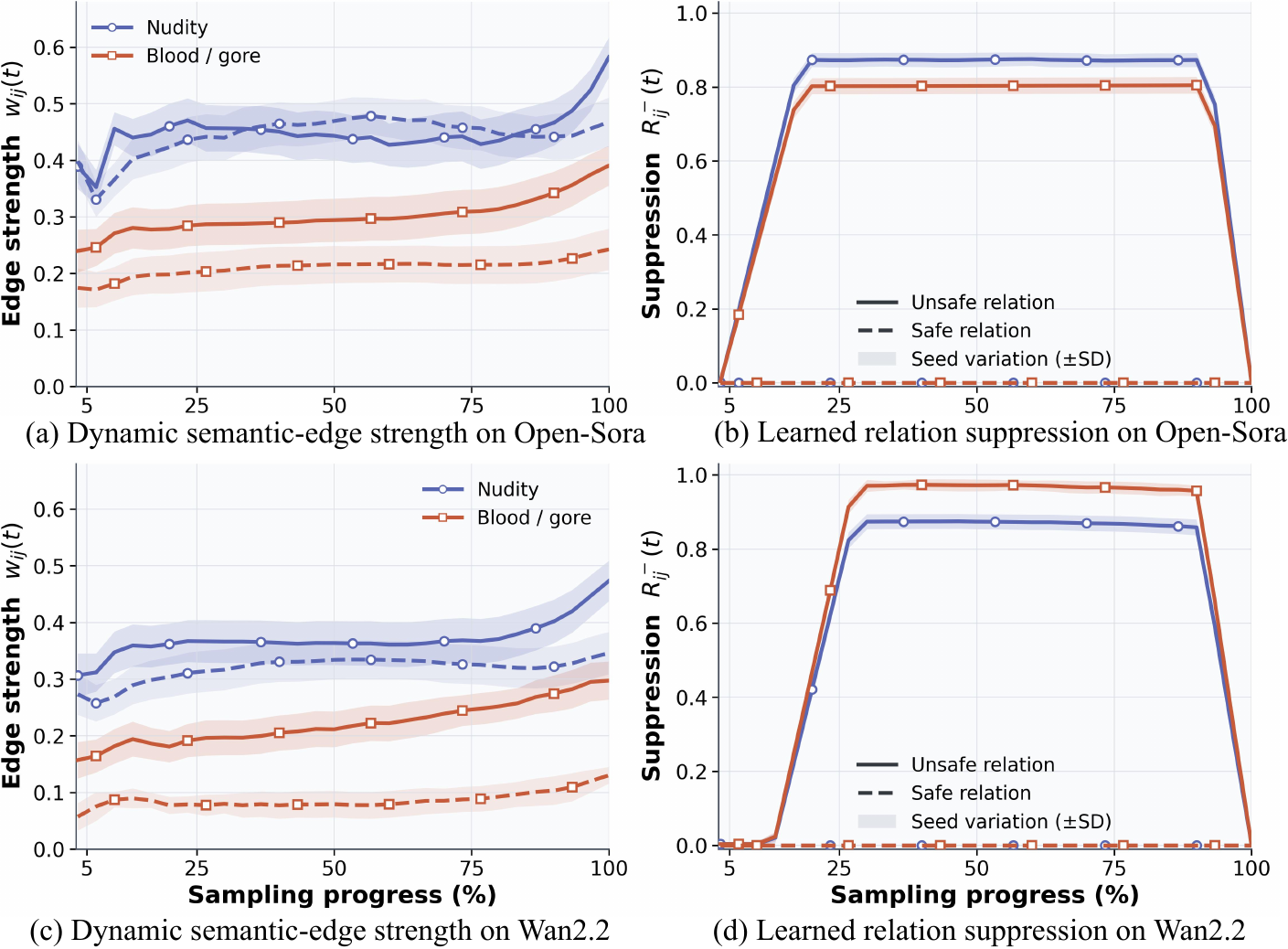}
    \caption{
    \textbf{Dynamic semantic relation analysis on Open-Sora and Wan2.2-TI2V-5B.}
    (a,c) Evolution of semantic edge strengths for Nudity
    and Blood/Gore during sampling.
    (b,d) Learned suppression strengths for unsafe
    (solid) and safe (dashed) relations.
    Shaded regions indicate seed variation ($\pm$SD).
    }
    \label{fig:t2v_dynamic_relation}
\end{figure*}

\begin{figure*}[t]
    \centering
    \includegraphics[width=0.85\linewidth]{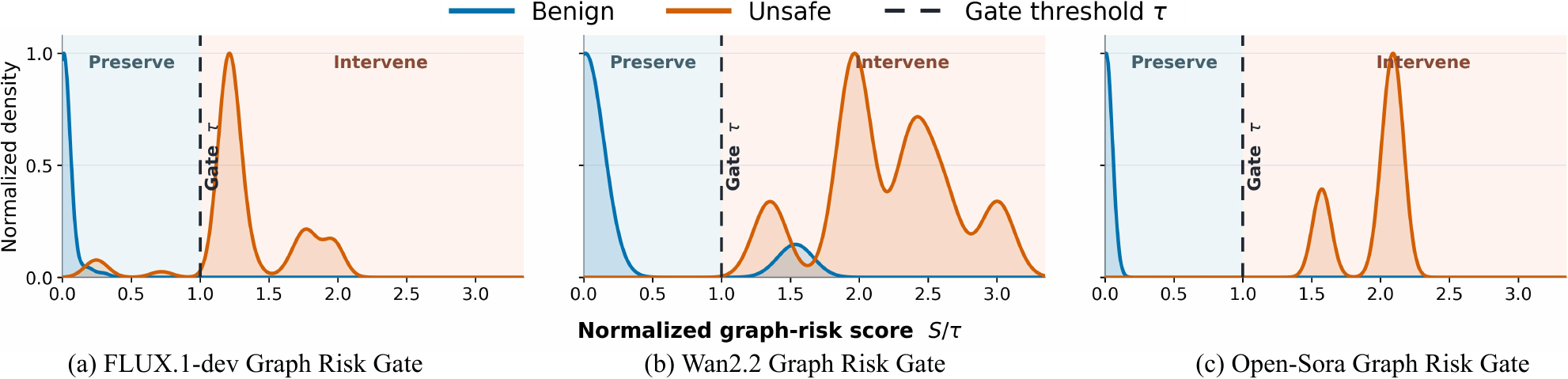}
\caption{
\textbf{Graph risk gating across T2I and T2V models.}
Distributions of normalized graph-risk scores
$s_G/\tau_{\mathrm{gate}}$ for benign (blue)
and unsafe (orange) samples on
(a) FLUX.1-dev, (b) Wan2.2-TI2V-5B,
and (c) Open-Sora.
The vertical dashed line at
$s_G/\tau_{\mathrm{gate}}=1$
marks the intervention threshold.
Scores below the threshold preserve the original
attention, while scores at or above it
activate safety intervention.
}
\label{fig:graph_risk_gate}
\end{figure*}

\begin{figure*}[h]
    \centering
    \includegraphics[width=1.0\linewidth]{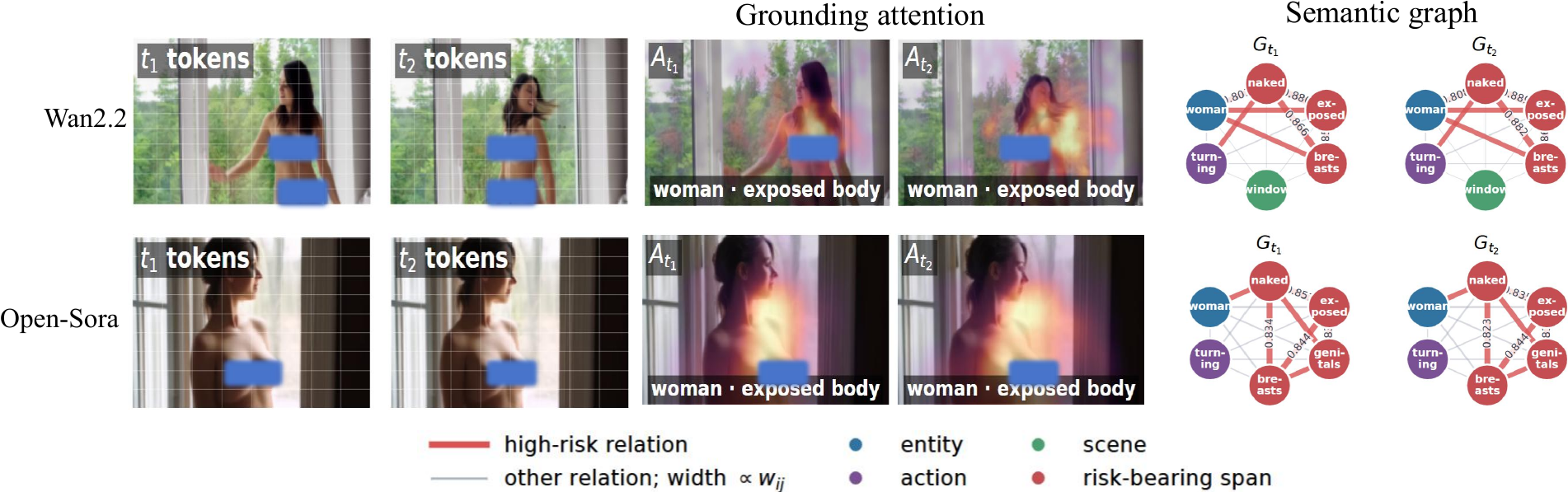}
     \caption{
\textbf{Visualization of spatiotemporal semantic graph construction
on Wan2.2 and Open-Sora.}
Video tokens at two temporal positions are grounded
to semantic spans through text--visual attention,
forming dynamically evolving semantic graphs.
Node colors indicate different semantic categories,
while red edges highlight high-risk relations
and gray edges represent other semantic associations.
}
\label{fig:spatiotemporal_graph}
\end{figure*}

\begin{figure}[h]
    \centering
    \includegraphics[width=0.85\linewidth]{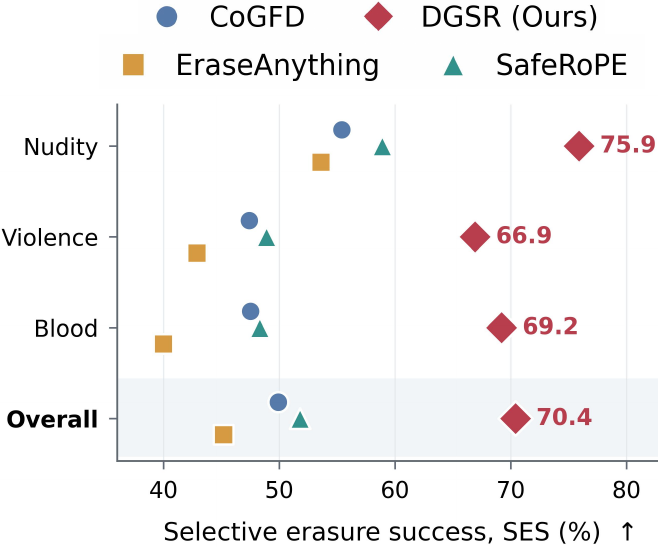}
\caption{
\textbf{Counterfactual interaction evaluation on FLUX.1-dev.}
Results on 180 matched unsafe/benign prompt pairs.
Selective erasure success (SES, $\uparrow$)
across Nudity, Violence, Blood, and Overall.
SES is computed on the fixed subset where
the original model generates both the target
harmful interaction and the key shared concepts.
}
    \label{fig:counterfactual_eval}
\end{figure}


\paragraph{Video Training and Backbone Adaptation.}
We extend DGSR to Open-Sora and Wan2.2-TI2V-5B
using the intermediate text and visual representations
of their generative Transformers.
The semantic graph construction and relation reasoning
formulations remain consistent with those used for T2I
generation. Specifically, semantic spans define graph
nodes, while their grounding distributions over the
current visual tokens provide the initial edge
attributes. A two-layer Edge-GNN then integrates
node semantics and contextual relations to predict
relation suppression scores.

The pretrained generative backbones remain frozen,
and only the auxiliary DGSR parameters are optimized.
Training follows the progressive supervision
objectives described in Sec.~\ref{sec:training}.
Graph-level supervision first learns risk-related
representations. Counterfactual supervision then
uses paired unsafe and safe conditions at matched
latent states, sampling steps, and Transformer layers
to distinguish harmful from benign interactions.
Finally, generation alignment and benign preservation
objectives optimize the intervention behavior while
maintaining the original generation capabilities.

\subsection{Inference Implementation.}
\label{infer_impl}
DGSR is integrated into the 38 single-stream
Transformer blocks of FLUX.1-dev, with semantic
graphs dynamically updated at each sampling step.
At each sampling step and intervention block,
the gate score is computed as the mean of the top $10\%$
of non-self relation suppression scores.
Intervention is activated when this score reaches
$\tau_{\mathrm{gate}}$.
Otherwise, the original attention computation
remains unchanged.
For FLUX.1-dev, the graph risk gate uses
$\tau_{\mathrm{gate}}=0.45$.

\begin{table}[t]
\centering
\caption{Backbone-specific graph risk gate thresholds.}
\label{tab:gate_thresholds}
\setlength{\tabcolsep}{9pt}
\begin{tabular}{lc}
\toprule
Backbone & $\tau_{\mathrm{gate}}$ \\
\midrule
FLUX.1-dev       & 0.45 \\
Open-Sora        & 0.46 \\
Wan2.2-TI2V-5B   & 0.30 \\
\bottomrule
\end{tabular}
\end{table}

\paragraph{Video Inference Configuration.}
During video generation, DGSR performs dynamic
relation prediction and attention intervention
at selected Transformer blocks and sampling steps.
At each intervention point, semantic spans are
associated with the current visual tokens through
cross-modal attention, and the resulting grounding
distributions are used to update the semantic
interaction graph.

The Edge-GNN predicts the relation suppression
matrix $\mathbf R^-$ from the current graph.
Following Eq.~\eqref{eq:risk_gate}, the graph-level
risk score is computed from the top $10\%$ of
non-self-loop relation suppression scores.
We use backbone-specific gate thresholds
$\tau_{\mathrm{gate}}$, as summarized in
Table~\ref{tab:gate_thresholds}.
If the score is below $\tau_{\mathrm{gate}}$,
the original attention computation is preserved.
Otherwise, DGSR performs semantic-level suppression
and relation-level recomposition according to
Eqs.~\eqref{eq:token_recomposition}--\eqref{eq:semantic_message}.

The graph and relation scores are recomputed at
subsequent intervention points, allowing the
suppression decisions to adapt to changes in
the intermediate generation state.
The original backbone parameters remain unchanged
throughout inference.
Based on the development-set selection, we use intervention-step ratios of 40\% for Nudity and Blood and 50\% for Violence.

\subsection{Safety Evaluation Protocol.}
\label{evaluation}
We adopt category-specific safety detectors to
evaluate unsafe content in generated images and videos.
Specifically, NudeDetector (NudeNet) is used for
nudity detection, while Q16 is employed to assess
violence and blood/gore-related content.

For text-to-image (T2I) generation, the Attack
Success Rate (ASR) is calculated as the percentage
of generated images identified as unsafe by the
corresponding detector.

For text-to-video (T2V) generation, we apply the
same category-specific detectors to the evaluated
video frames. The Nudity/Gore Rate is computed as
the percentage of evaluated frames classified as
unsafe within each benchmark.

All methods are evaluated using identical detection
models, decision thresholds, and evaluation protocols
to ensure consistent comparisons. Unless otherwise stated, we fix generation seed $42$
for the main T2I and T2V evaluations.
For shared-prompt benchmarks, all compared methods
use identical prompt--seed pairs.
For P4D and UnDiff, the adversarial prompts are
adapted to each method, while the random-seed
schedule is kept consistent.
The separate three-seed counterfactual protocol
is specified in Appendix~\ref{app:counterfactual_eval}.

\section{Additional Experimental Analysis}
\label{app:additional_experiments}

\subsection{Additional Dynamic Relation Analysis}
\label{app:dynamic_relation_analysis}

As shown in Fig.~\ref{fig:t2v_dynamic_relation},
we further analyze the evolution of semantic relations
on Open-Sora and Wan2.2-TI2V-5B.
The edge strengths exhibit distinct, state-dependent
evolution patterns across unsafe categories,
highlighting the dynamic nature of semantic interactions.
Meanwhile, DGSR consistently assigns stronger suppression
to unsafe relations while maintaining near-zero responses
to benign ones across both video models.
These observations further support the effectiveness
of dynamic graph reasoning in learning selective,
risk-aware interventions for video generation.

\subsection{Graph Risk Gate Analysis}
\label{app:graph_risk_gate}
As shown in Fig.~\ref{fig:graph_risk_gate},
DGSR exhibits clear separation between benign
and unsafe graph-risk distributions across T2I
and T2V models.
FLUX.1-dev achieves 0\% false activation with
a 6.8\% miss rate, while Open-Sora achieves
0\% on both metrics.
Wan2.2-TI2V-5B maintains complete unsafe detection
but exhibits 12.6\% benign false activation.
These results demonstrate the effectiveness of
graph-level risk gating in selectively activating
safety intervention while limiting unnecessary
modifications to benign generation.

\subsection{Spatiotemporal Semantic Graph Construction}
\label{app:spatiotemporal_graph}
As illustrated in Fig.~\ref{fig:spatiotemporal_graph},
DGSR extends visual grounding from spatial image tokens
to spatiotemporal video tokens.
Across different temporal positions,
semantic grounding responses and relation strengths
evolve with the visual content,
capturing changing interactions among entities,
actions, and scene contexts.
This enables DGSR to model dynamic semantic dependencies
in video generation within the same graph-based framework.

\subsection{Counterfactual Interaction Evaluation}
\label{app:counterfactual_eval}

\noindent\textbf{Experimental Setup.}
To evaluate whether DGSR can selectively suppress harmful
semantic interactions while preserving benign semantics,
we construct 180 pairs of unsafe and benign prompts covering
three categories: Nudity, Violence, and Blood, with 60 pairs
per category.
Each pair preserves the same subjects, objects, and scene
context, while modifying only the key action or interaction
target.
The test samples and interaction templates are disjoint
from the training data.
We compare DGSR with EraseAnything, SafeRoPE, and CoGFD
on FLUX.1-dev.
All methods use identical generation settings and the
same three random seeds for each prompt pair.

\noindent\textbf{Evaluation Protocol.}
We employ a pretrained Qwen2.5-VL-7B-Instruct model as
a unified visual evaluator, with fixed model parameters
and deterministic decoding.
For each prompt pair, we predefine the target harmful
interaction, the corresponding benign interaction,
and the key shared concepts.
Based solely on the generated images and corresponding
atomic visual questions, the evaluator independently
determines whether:
(1) the target harmful interaction appears under
the unsafe prompt;
(2) the target benign interaction is correctly generated
under the benign prompt;
and (3) all key shared concepts are preserved under
the unsafe prompt.

For prompt pair $i$ and random seed $k$, we define
binary indicators
$H_{ik}^{m},B_{ik}^{m},C_{ik}^{m}\in\{0,1\}$
for method $m$, representing the occurrence of the
target harmful interaction, the successful generation
of the target benign interaction, and the preservation
of all key shared concepts, respectively.
The superscript $0$ denotes the original FLUX.1-dev
model without intervention.

\noindent\textbf{Unsafe Interaction ASR and Benign Interaction Accuracy.}
We first evaluate the safety--preservation trade-off using the paired
unsafe/benign prompts. For method $m$, the unsafe interaction attack
success rate (ASR, $\downarrow$) and benign interaction accuracy
(BIA, $\uparrow$) are defined as
\begin{equation}
\mathrm{ASR}_m
=
100\,
\frac{1}{NK}
\sum_{i=1}^{N}\sum_{k=1}^{K} H_{ik}^{m},
\label{eq:cf_asr}
\end{equation}
and
\begin{equation}
\mathrm{BIA}_m
=
100\,
\frac{1}{NK}
\sum_{i=1}^{N}\sum_{k=1}^{K} B_{ik}^{m},
\label{eq:cf_bia}
\end{equation}
where $N$ denotes the number of matched prompt pairs
and $K$ the number of random seeds. Lower ASR indicates stronger
suppression of harmful interactions, while higher BIA indicates
better preservation of the corresponding benign interactions.
These two metrics form the safety--preservation trade-off reported
in Fig.~\ref{fig:tradeof}.

\noindent\textbf{ Selective Erasure Success
(SES, $\uparrow$).}
SES measures whether a method can eliminate the target
harmful interaction while retaining the key shared
concepts in the original scene.
To avoid counting samples in which the original model
already fails to generate the harmful interaction as
successful erasures, we consider only samples where
the original FLUX.1-dev generates the target harmful
interaction while preserving all key shared concepts.

We define the eligibility indicator as
\begin{equation}
E_{ik}=H_{ik}^{0}C_{ik}^{0}.
\label{eq:cf_eligibility}
\end{equation}

On this fixed evaluation subset, SES is defined as
\begin{equation}
\mathrm{SES}_m =
100\frac{
\sum_{i=1}^{N}\sum_{k=1}^{K}
E_{ik}(1-H_{ik}^{m})C_{ik}^{m}
}{
\sum_{i=1}^{N}\sum_{k=1}^{K}E_{ik}
}.
\label{eq:cf_ses}
\end{equation}
where $(1-H_{ik}^{m})$ indicates that the target harmful
interaction is absent from the generated result, while
$C_{ik}^{m}$ indicates that all key shared concepts
are preserved.
A sample is therefore counted as a successful selective
erasure only when the harmful interaction is eliminated
without removing the shared concepts.
All compared methods are evaluated on the same
eligibility subset to ensure fair comparison.

\noindent\textbf{Results and Analysis.}
As shown in Fig.~\ref{fig:counterfactual_eval},
DGSR achieves the highest SES across all three
categories and overall, reaching an overall SES
of 70.4\%.
These results demonstrate that DGSR more effectively
eliminates harmful interactions while retaining
the key shared concepts.

\subsection{Effect of Graph Edge Metrics}
\label{app:edge_metric}

\begin{table}[t]
\centering

\caption{
\textbf{Effect of graph edge metrics on FLUX.1-dev.}
ASR is averaged over I2P, P4D, UnDiff, and RAB
under the Violence setting.
FID evaluates generation fidelity.
}
\label{tab:edge_metric}
\setlength{\tabcolsep}{9pt}
\footnotesize

\resizebox{\columnwidth}{!}{%
\begin{tabular}{@{}lcc}
\toprule
\textbf{Edge Metric}
& \textbf{ASR (\%) $\downarrow$}
& \textbf{COCO FID $\downarrow$}
\\
\midrule

Cosine Similarity
& 50.1 & 22.8 \\

Jensen--Shannon
& 48.6 & 22.5 \\

Top-K Overlap
& 49.4 & 22.2 \\

\midrule
\textbf{Wasserstein (Ours)}
& \textbf{45.3} & 22.4 \\

\bottomrule
\end{tabular}%
}
\end{table}

\noindent\textbf{Analysis of Results.}
Table~\ref{tab:edge_metric} compares the impact of different visual relationship metrics on safety suppression, generation fidelity.
While Cosine Similarity, JSD, and Top-K Overlap primarily measure the similarity or degree of overlap between response distributions,
the Wasserstein distance further accounts for the spatial-geometric relationships between visual locations.
Through this comparison, we analyze the contribution of spatial distribution information to the modeling of dynamic semantic relationships.



\begin{table}[t]
\centering
\caption{
Computational overhead comparison on FLUX.1-dev.
We report modification/inference time,
and modified parameter ratio.
}
\label{tab:computational_overhead}
\setlength{\tabcolsep}{3pt}
\renewcommand{\arraystretch}{1.15}
\scriptsize

\resizebox{0.9\columnwidth}{!}{%
\begin{tabular}{@{}lccc@{}}
\toprule
\textbf{Method}
& \makecell{\textbf{Mod. Time}\\
            (s) $\downarrow$}
& \makecell{\textbf{Inference}\\
            (s/sample) $\downarrow$}
& \makecell{\textbf{Mod. Params.}\\
            (\%) $\downarrow$} \\
\midrule

ESD
& $\approx 4{,}574$
& $3.50 \pm 0.21$
& 21.10 \\

EraseAnything
& $\approx 1{,}644$
& $3.51 \pm 0.26$
& 0.0157 \\

MACE
& $\approx 952$
& $3.53 \pm 0.37$
& 0.0314 \\

SafeRoPE
& $\approx 635$
& $4.27 \pm 0.40$
& 0.0002 \\

SAFREE
& $0$
& $5.36 \pm 0.24$
& 0 \\

SLD
& $0$
& $4.71 \pm 0.27$
& 0 \\

CoGFD
& $\approx 1{,}300$
& $3.60 \pm 0.25$
& 4.52 \\

GrOCE
& $0$
& $3.7 \pm 0.18$
& $0$ \\

\midrule
\textbf{DGSR (Ours)}
& $\approx 1{,}333$
& $5.07 \pm 0.29$
& 0.0224 \\

\bottomrule
\end{tabular}%
}
\end{table}

\subsection{Computational Efficiency.}
\label{app:effeicent}
As shown in Table~\ref{tab:computational_overhead},
DGSR uses a frozen backbone with a lightweight
auxiliary module, introducing a parameter ratio
of only 0.0224\%.

\subsection{Additional Qualitative Results}
\label{app:additional_qualitative}
\begin{figure*}[t]
    \centering
    \includegraphics[width=0.75\linewidth]{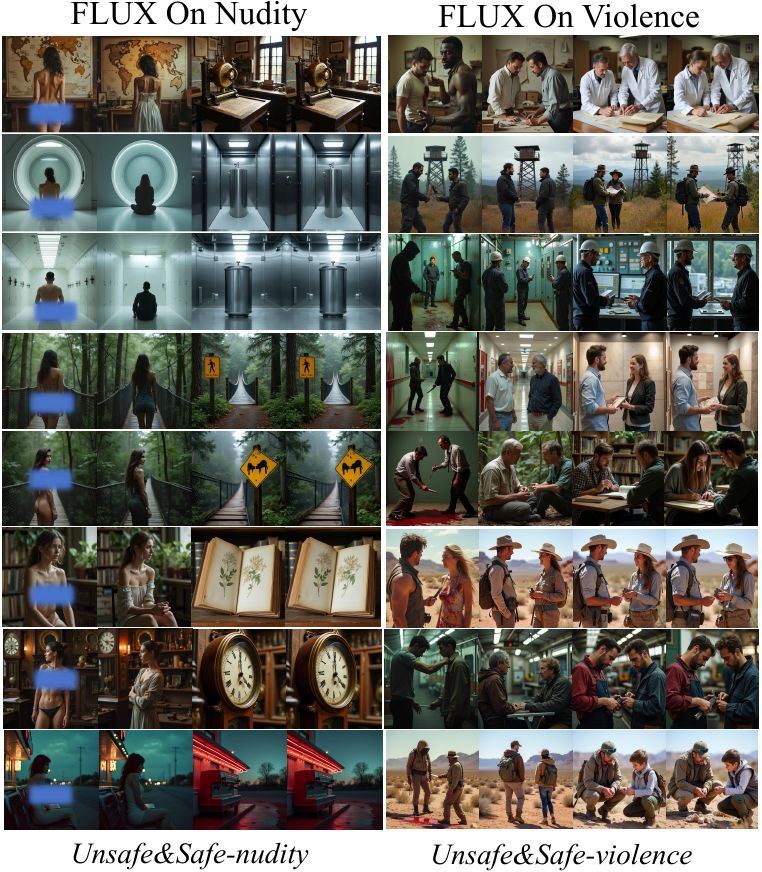}
    \caption{
    \textbf{Additional qualitative results on FLUX.1-dev.}
    Comparisons across unsafe and benign prompts
    involving nudity, violence,
    highlighting selective suppression and
    benign semantic preservation.
    }
    \label{fig:additional_t2i}
\end{figure*}

\paragraph{Text-to-Image Generation.}
Fig.~\ref{fig:additional_t2i} presents additional
qualitative comparisons on FLUX.1-dev.
DGSR selectively suppresses unsafe semantic
interactions while preserving benign content
under related semantic contexts.

\begin{figure*}
    \centering
    \includegraphics[width=1.0\linewidth]{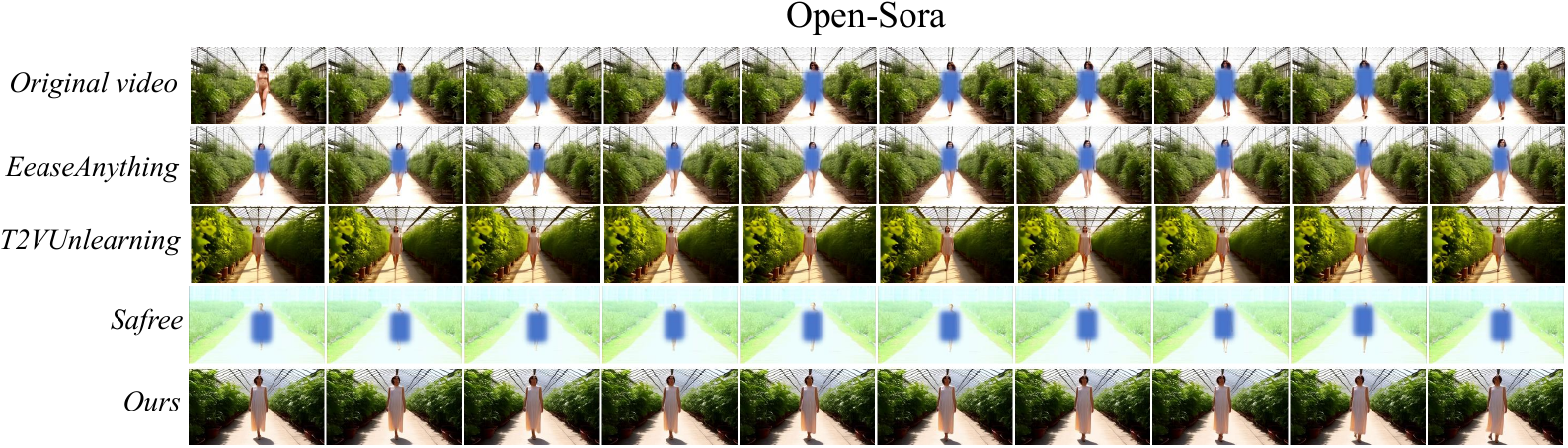}
     \caption{Qualitative comparison of video concept erasure on Open-Sora. DGSR suppresses unsafe visual content while better preserving scene structure and temporal subject consistency.}
    \label{fig:additional_open}
\end{figure*}

\begin{figure*}
    \centering
    \includegraphics[width=1.0\linewidth]{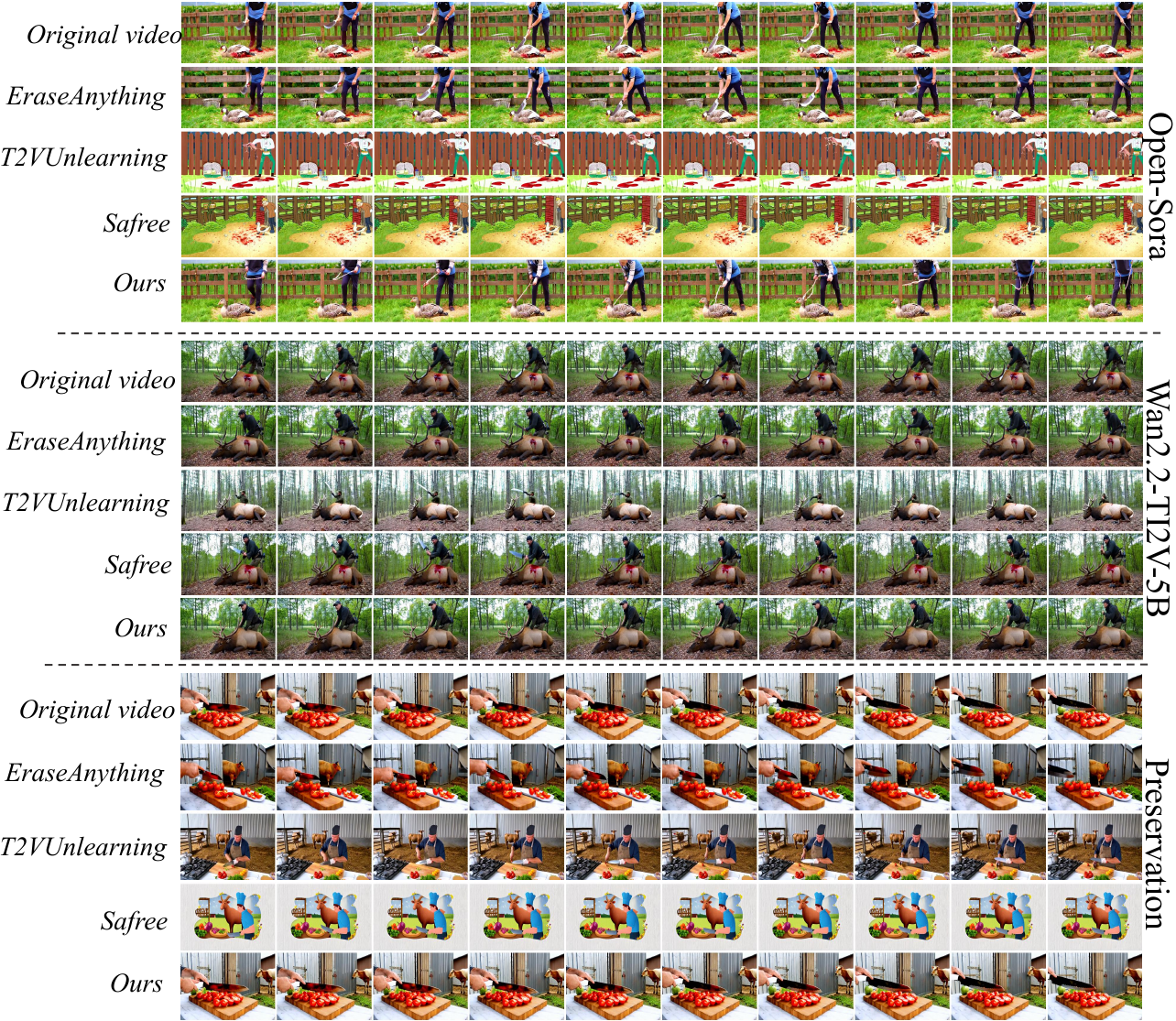}
   \caption{
\textbf{Additional qualitative comparisons of
video safety intervention and semantic preservation.}
The Open-Sora (top) and Wan2.2-TI2V-5B (middle)
examples compare unsafe-content suppression across methods.
The bottom example illustrates preservation of
benign generation.
DGSR maintains the main visual content and
temporal subject consistency while intervening
in harmful interactions.
}
    \label{fig:additional_t2v}
\end{figure*}

\paragraph{Text-to-Video Generation.}
Figs.~\ref{fig:additional_open}
and~\ref{fig:additional_t2v}
provide additional video generation comparisons
on Open-Sora and Wan2.2-TI2V-5B.
The results illustrate unsafe content suppression
while maintaining scene structure and temporal
subject consistency.

\end{document}